\documentclass[journal]{IEEEtran}

\usepackage[mathscr]{eucal}
\usepackage[cmex10]{amsmath}
\usepackage{epsfig,epsf,psfrag}
\usepackage{amssymb,amsmath,amsthm,amsfonts,latexsym}
\usepackage{amsmath,graphicx,bm,xcolor,url}
\usepackage[caption=false]{subfig} 
\usepackage{fixltx2e}%ordering of single and double column floats
\usepackage{array}%array and tabular environments
\usepackage{verbatim}
\usepackage{bm}
\usepackage{algorithmic, cite}
\usepackage{algorithm}
\usepackage{verbatim}
\usepackage{textcomp}
\usepackage{mathrsfs}
\usepackage{epstopdf}
\catcode`~=11 \def\UrlSpecials{\do\~{\kern -.15em\lower .7ex\hbox{~}\kern .04em}} \catcode`~=13 

\allowdisplaybreaks[3]

\newcommand{\calI}{\mathcal{I}}

\newcommand{\calN}{\mathcal{N}}
\newcommand{\calO}{\mathcal{O}}

\newcommand{\calS}{\mathcal{S}}

\newcommand{\calU}{\mathcal{U}}

\newcommand{\ba}{\mathbf{a}}
\newcommand{\bA}{\mathbf{A}}
\newcommand{\bb}{\mathbf{b}}
\newcommand{\bB}{\mathbf{B}}
\newcommand{\bc}{\mathbf{c}}
\newcommand{\bC}{\mathbf{C}}
\newcommand{\bd}{\mathbf{d}}
\newcommand{\bD}{\mathbf{D}}
\newcommand{\be}{\mathbf{e}}
\newcommand{\bE}{\mathbf{E}}

\newcommand{\bF}{\mathbf{F}}
\newcommand{\bg}{\mathbf{g}}
\newcommand{\bG}{\mathbf{G}}

\newcommand{\bH}{\mathbf{H}}

\newcommand{\bI}{\mathbf{I}}

\newcommand{\bL}{\mathbf{L}}

\newcommand{\bM}{\mathbf{M}}

\newcommand{\bp}{\mathbf{p}}

\newcommand{\bq}{\mathbf{q}}
\newcommand{\bQ}{\mathbf{Q}}

\newcommand{\bt}{\mathbf{t}}
\newcommand{\bT}{\mathbf{T}}
\newcommand{\bu}{\mathbf{u}}
\newcommand{\bU}{\mathbf{U}}

\newcommand{\bx}{\mathbf{x}}
\newcommand{\bX}{\mathbf{X}}

\newcommand{\bY}{\mathbf{Y}}

\DeclareMathAlphabet{\mathbsf}{OT1}{cmss}{bx}{n}
\DeclareMathAlphabet{\mathssf}{OT1}{cmss}{m}{sl}% slanted sans serif

\DeclareSymbolFont{bsfletters}{OT1}{cmss}{bx}{n}  
\DeclareSymbolFont{ssfletters}{OT1}{cmss}{m}{n}
\DeclareMathSymbol{\bsfGamma}{0}{bsfletters}{'000}
\DeclareMathSymbol{\ssfGamma}{0}{ssfletters}{'000}
\DeclareMathSymbol{\bsfDelta}{0}{bsfletters}{'001}
\DeclareMathSymbol{\ssfDelta}{0}{ssfletters}{'001}
\DeclareMathSymbol{\bsfTheta}{0}{bsfletters}{'002}
\DeclareMathSymbol{\ssfTheta}{0}{ssfletters}{'002}
\DeclareMathSymbol{\bsfLambda}{0}{bsfletters}{'003}
\DeclareMathSymbol{\ssfLambda}{0}{ssfletters}{'003}
\DeclareMathSymbol{\bsfXi}{0}{bsfletters}{'004}
\DeclareMathSymbol{\ssfXi}{0}{ssfletters}{'004}
\DeclareMathSymbol{\bsfPi}{0}{bsfletters}{'005}
\DeclareMathSymbol{\ssfPi}{0}{ssfletters}{'005}
\DeclareMathSymbol{\bsfSigma}{0}{bsfletters}{'006}
\DeclareMathSymbol{\ssfSigma}{0}{ssfletters}{'006}
\DeclareMathSymbol{\bsfUpsilon}{0}{bsfletters}{'007}
\DeclareMathSymbol{\ssfUpsilon}{0}{ssfletters}{'007}
\DeclareMathSymbol{\bsfPhi}{0}{bsfletters}{'010}
\DeclareMathSymbol{\ssfPhi}{0}{ssfletters}{'010}
\DeclareMathSymbol{\bsfPsi}{0}{bsfletters}{'011}
\DeclareMathSymbol{\ssfPsi}{0}{ssfletters}{'011}
\DeclareMathSymbol{\bsfOmega}{0}{bsfletters}{'012}
\DeclareMathSymbol{\ssfOmega}{0}{ssfletters}{'012}

\DeclareMathOperator*{\argmax}{arg\,max}

\DeclareMathOperator{\col}{col}

\newtheorem{theorem}{Theorem} 
\newtheorem{lemma}[theorem]{Lemma}

\newtheorem{corollary}[theorem]{Corollary}
\newtheorem{definition}{Definition}

\newtheorem{remark}{Remark}

\newtheorem{assumption}{Assumption}
\newtheorem{data model}{Data Model}

\newcommand{\qednew}{\nobreak \ifvmode \relax \else
      \ifdim\lastskip<1.5em \hskip-\lastskip
      \hskip1.5em plus0em minus0.5em \fi \nobreak
      \vrule height0.75em width0.5em depth0.25em\fi}

\begin{document}
\title{Coherence Pursuit: Fast, Simple, and Robust Principal Component Analysis}

\author{Mostafa~Rahmani, \IEEEmembership{Student Member,~IEEE} and George~K.~Atia,~\IEEEmembership{Member,~IEEE} % <-this % stops a space
\thanks{This work was supported in part by NSF CAREER Award CCF-1552497 and NSF Grant CCF-1320547.

The authors are with the Department of Electrical and Computer Engineering, University of Central Florida, Orlando, FL 32816 USA (e-mails: mostafa@knights.ucf.edu, george.atia@ucf.edu).}% <-this % stops a spa
}

\markboth{}%
%\markboth{Journal of \LaTeX\ Class Files,~Vol.~11, No.~4, %December~2012}%
{Shell \MakeLowercase{\textit{et al.}}: Bare Demo of IEEEtran.cls for Journals}
\maketitle

\begin{abstract}

This paper presents a remarkably simple, yet powerful, algorithm termed Coherence Pursuit (CoP) to robust Principal Component Analysis (PCA).
% In this paper, a super simple but  strong robust Principal Component Analysis (PCA) algorithm is presented.
% In the proposed method, we distinguish an outlying data point from an inlier by comparing their coherence with the rest of the data points.
As inliers lie in a low-dimensional subspace and are mostly correlated, an inlier is likely to have strong mutual coherence with a large number of data points. By contrast, outliers either do not admit low dimensional structures or form small clusters. In either case, an outlier is unlikely to bear strong resemblance to a large number of data points. Given that, CoP sets an outlier apart from an inlier by comparing their coherence with the rest of the data points.
%hence if there are enough inliers it is highly likely that if there are enough number of inliers, the inliers have strong mutual coherence together. However, outliers mostly do not follow a low dimensional structure and there are not too many outlying data points which are  similar to each other.
% We study all the mutual coherences with a simple matrix multiplication.
The mutual coherences are computed by forming the Gram matrix of the normalized data points. Subsequently, the sought subspace is recovered from the span of the subset of the data points that exhibit strong coherence with the rest of the data. As CoP only involves one simple matrix multiplication, it is significantly faster than the state-of-the-art robust PCA algorithms. We derive analytical performance guarantees for CoP under different models for the distributions of inliers and outliers in both noise-free and noisy settings. % CoP is shown to be robust to both unstructured and structured outliers.  %   by vast majority of the data outlier.
% Roughly, the proposed method involves only one matrix multiplication and it makes Coherence pursuit significantly faster than the state of the art robust PCA algorithms.
% To the best of our knowledge, this
CoP is the first robust PCA algorithm that is simultaneously non-iterative, provably robust to both unstructured and structured outliers, and can tolerate a large number of unstructured outliers.
% the proposed method is the first provable robust PCA algorithm which is not iterative, can tolerate high amount of outliers and is even robust to linearly dependent outliers.
 % We analysis the proposed algorithm with a random model for the distribution of the inliers/outliers and it is shown that the proposed method can recover the correct subspace even if the vast majority of the data is outlier.
\end{abstract}


\IEEEpeerreviewmaketitle

%%%%%%%%% BODY TEXT
\section{Introduction}

%Principal Component Analysis (PCA) is a key mathematical tool used in a broad range of applications. PCA aims to approximate the data with a low-dimensional subspace.

Standard tools such as Principal Component Analysis (PCA) have been instrumental in reducing dimensionality by finding linear projections of high-dimensional data along the directions where the data is most spread out to minimize information loss. These techniques are widely applicable in
% Such linear models are highly pertinent to
a broad range of data analysis problems, including problems in computer vision, image processing, machine learning and bioinformatics \cite{lamport44,lamport45,dadaneh2016bayesian,taalimi2015multimodal,hosseini2014support,rahmani2015innovation}.

Given a data matrix $\bD \in \mathbb{R}^{m \times n}$, PCA finds an $r$-dimensional subspace by solving
%. Therefore, the conventional PCA problem can be written as
\begin{eqnarray}
\underset{\hat{\bU}}{\min} \| \bD - \hat{\bU} \hat{\bU}^T \bD \|_F \quad \text{subject to} \quad \hat{\bU}^T \hat{\bU} = \bI,
\label{eq1}
\end{eqnarray}
where $\hat{\bU} \in \mathbb{R}^{m \times r}$ is an orthonormal basis for the $r$-dimensional subspace, $\bI$ denotes the identity matrix and $\|.
\|_F$ the Frobenius norm. Despite its notable impact on exploratory data analysis and multivariate analyses, PCA is notoriously sensitive to outliers that prevail much of the real world data since the solution to (\ref{eq1}) can arbitrarily deviate from the true subspace in  presence of a small number of outlying data points that do not conform with the low-dimensional model
\cite{hauberg2015scalablen,nie2014optimal,luo2016avoiding,nie2015joint,nie2011robust,amini2014outlier,lerman2014l_p,maronna2005principal,mccoy2011two,xu2010principal,zhang2012robust}.

%While PCA is useful when the data has low intrinsic dimension, it is notoriously sensitive to outliers in the sense that the solution to % the least-square method (\ref{eq1}) can arbitrarily deviate from the true underlying subspace if a small portion of the data is not contained in this low-dimensional subspace.

%The least-squares method is notoriously sensitive to outliers in the sense that the output of PCA can arbitrarily deviate be dramatically impacted even if just a small portion of the data does not follow the low-dimensional model.

% As outliers prevail much of the real data, a large body of research has focused on developing robust PCA algorithms that are not unduly affected by the presence of outliers.
As a result, much research work was devoted to investigate PCA algorithms that are robust to outliers.
The corrupted data can be expressed as
\begin{eqnarray}
\bD = \bL + \bC \: ,
\label{eq2}
\end{eqnarray}
where $\bL$ is a low rank matrix whose columns span a low-dimensional subspace, and the matrix $\bC$ models the data corruption, and is referred to as the outlier matrix.
%the corruption from the low rank matrix. We call $\bC$ the outlier matrix.
Two main models for the outlier matrix were considered in the literature -- these two models are mostly incomparable in theory, practice and analysis techniques. The first corruption model is the element-wise model in which $\bC$ is a sparse matrix with arbitrary support, whose entries can have arbitrarily large magnitudes \cite{rahmani2016subspace,bouwmans2017decomposition,rahmani2015highn,lamport1,lamport22,netrapalli2014non,yi2016fast}. In view of the arbitrary support of $\bC$, any of the columns of $\bL$ may be affected by the non-zero elements of $\bC$. We do not consider this model in this paper. The second model, which is the focus of our paper, is a column-wise model wherein only a fraction of the columns of $\bC$ are non-zero, wherefore a portion of the columns of $\bL$ (the so-called inliers) remain unaffected by $\bC$ \cite{lerman2014fast,lamport10,feng2016robust,lamport24,rahmani2015randomized}.
%
%Two main models for data corruption that are in fact incomparable for the most part were considered in the literature, namely, element-wise and column-wise corruption. In the former model, $\bC$ is an element-wise sparse matrix with arbitrary support, whose entries can have arbitrarily large magnitudes \cite{rahmani2016subspace,lamport1,lamport22,netrapalli2014non,yi2016fast}. In this model, all the columns of $\bL$ may be affected by the non-zero elements of $\bC$ given its arbitrary support pattern.
%There are two main models for the corruption, element-wise corruption and column-wise corruption.
%In the element-wise model, the matrix $\bC$ is an element-wise sparse matrix. This means that a small portion of the elements of $\bC$ are non-zero, but we cannot assume any structure on the pattern of the non-zero elements \cite{lamport1}. In this model, all the columns of $\bL$ can be affected by the non-zero elements of the outlier matrix $\bC$.
%
%In the column-wise model, a portion of the columns of $\bC$ are non-zero and these non-zero columns do not lie in the column space of $\bL$ \cite{lamport10 , lamport24,myconf,rahmani2015randomized}. Thus, a portion of the columns of $\bL$, the so-called inliers, are unaffected by $\bC$. Suppose matrix $\bB$ contains the non-zero columns of $\bC$ and $\bA$ contains the inliers.

% This paper focuses on the column-wise outlier model according to the following data model.

\subsection{The inlier-outlier structure}
We formally describe the data model adopted in this paper, which only focuses on the column-wise outlier model.
\begin{data model}
The given data matrix $\bD$ satisfies the following.\\
1. The matrix $\bD$ can be expressed as \begin{eqnarray}
\bD = \bL + \bC = [\bA \: \: \bB] \: \bT \:,
\label{eq:data_model}
\end{eqnarray}
where $\bA \in \mathbb{R}^{m \times n_1}$, $\bB \in \mathbb{R}^{m \times n_2}$, and $\bT$ is an arbitrary permutation matrix. \\
2. The columns of $\bA$ lie in an $r$-dimensional subspace $\calU = \col(\bL)$, the column space of $\bL$. The columns of $\bB$ do not lie entirely in $\calU$, i.e., the $n_1$ columns of $\bA$ are the inliers and the $n_2$ columns of $\bB$ are the outliers.
\end{data model}

We consider two types of column-wise outliers. The first type consists of data points which do not follow a low-dimensional structure. In addition, a small number of these points are not linearly dependent. We refer to this type as `unstructured outliers'.
Unstructured outliers are typically modeled as data points scattered uniformly at random in the ambient space
\cite{heckel2013robust,tsakiris2015dual,soltanolkotabi2012geometric}.
%Whereas unstructured outliers are generally distinguishable even if they dominate the data \cite{heckel2013robust,soltanolkotabi2012geometric}, the problem becomes more challenging when the data is highly noisy or corrupted. Unstructured outliers are generally distinguishable even if they dominate
Such outliers are generally distinguishable even if they dominate the data \cite{heckel2013robust,soltanolkotabi2012geometric}. A conceivable scenario for unstructured outliers is when a set of the data points are
intolerably noisy or highly corrupted.
The second type, which we refer to as `structured outliers', concerns data points which are linearly dependent or form a cluster. In other words, structured outliers exist in small numbers relative to the size of the data but form some low-dimensional structure different from that of most of the data points.
% Obviously, the population of these outliers should be small comparing to size of data.
Structured outliers are often associated with rare patterns or events of interest, such as important regions in an image \cite{lamport9}, malignant tissues \cite{karrila2011comparison}, or web attacks \cite{kruegel2003anomaly}.
% This type of outlier mostly corresponds to the rare events which  detecting them is of particular interest  in many applications. Detecting these rare events could  correspond to finding the important regions in an image \cite{lamport9}, identifying malignant tissues \cite{karrila2011comparison}, or  detecting web attacks \cite{kruegel2003anomaly}.

The column-wise model for robust PCA has direct bearing on a host of applications in signal processing and machine learning, which % a well known and important research problem in both signal processing and machine learning communities.
spurred enormous progress in dealing with subspace recovery in the presence of outliers. This paper is motivated by some of the limitations of existing techniques, which we further detail in Section \ref{sec:rel_work} on related work.
\textcolor{black}{
The vast majority of existing approaches to robust PCA have \emph{high computational complexity}, which makes them unsuitable in high-dimensional settings}. For instance, many of the existing iterative techniques incur a long run time as they require a large number of iterations, each with a Singular Value Decomposition (SVD) operation.
% Many approaches have been proposed to address this problem but most of them require too many high complexity iterations.
% Mostly, each iteration involves one SVD operation which imposes high computation complexity and long running time for high dimensional data.
% In addition, mostly the exact subspace recovery can not be guaranteed with  the iterative solvers.
Also, most iterative solvers have \emph{no provable guarantees} for exact subspace recovery.
Moreover, some of the existing methods rest upon restrictive definitions of outliers. For instance, \cite{heckel2013robust,tsakiris2015dual,soltanolkotabi2012geometric} can only detect unstructured randomly distributed outliers and \cite{lamport10} requires $\bC$ to be column sparse.
In this paper, we present a new provable non-iterative robust PCA algorithm, dubbed Coherence Pursuit (CoP), which involves one simple matrix multiplication, and thereby achieves remarkable \textcolor{black}{speedups} over the state-of-the-art algorithms.
% This makes the proposed method remarkably faster than the state of the art algorithms.
CoP does not presume a restrictive model for outliers and provably detects both structured and unstructured outliers.
In addition, it can tolerate a large number of unstructured outliers -- even if the ratio of inliers to outliers $\frac{n_1}{n_2}$ approaches zero.

\subsection{Notation and definitions}
Bold-face upper-case and lower-case letters are used to denote matrices and vectors, respectively. Given a matrix $\bA$, $\| \bA \|$ denotes its spectral norm, $\| \bA \|_{*}$ its nuclear norm, and $\col(\bA)$ its column space. For a vector $\ba$, $\| \ba \|_p$ denotes its $\ell_p$-norm and $\ba(i)$ its $i^{\text{th}}$ element. Given two matrices $\bA_1$ and $\bA_2$ with an equal number of rows, the matrix
\[
\bA_3 = [\bA_1 \: \: \bA_2]
\]
is the matrix formed by concatenating their columns. For a matrix $\bA$, $\ba_i$ denotes its $i^{\text{th}}$ column, and $\bA_{-i}$ is equal to $\bA$ with the $i^{\text{th}}$ column removed.
Given matrix $\bA$,  $\| \bA \|_{1,2} = \sum_{i} \| \ba_i \|_2$.
The function orth$(\cdot)$ % is defined similar to the function orth$(\cdot)$ in MATLAB, which
returns an orthonormal basis for the range of its matrix argument. In addition, $\mathbb{S}^{m-1}$ denotes the unit $\ell_2$-norm sphere in $\mathbb{R}^m$.

\begin{definition}
The coherence value corresponding to the $i^{\text{th}}$ data point with parameter $p$ is defined as
$$
\bp(i) = \sum_{k = 1 \atop k \neq i}^{n} |\bx_i^T \bx_k|^p\:,
$$
where $\bx_j = \bd_j / \| \bd_j \|_2$ for $ 1\leq j \leq n$. The vector $\bp \in \mathbb{R}^n$ contains the coherence values for all the data points.
\label{def:coh}
\end{definition}

\section{Related Work}
\label{sec:rel_work}
Some of the earliest approaches to robust PCA relied on robust estimation of the data covariance matrix, such as S-estimators, the minimum covariance determinant,
the minimum volume ellipsoid, and the Stahel-Donoho estimator \cite{lamport47}. This is a class of iterative approaches that compute a full SVD or eigenvalue decomposition in each iteration and generally have no explicit performance guarantees. % they are not supported with explicit performance guarantee.
The performance of these approaches greatly degrades when $\frac{n_1}{n_2} \le 0.5$.

To enhance robustness to outliers, another approach is to replace the Frobenius norm in (\ref{eq1}) with other norms \cite{lamport18}. For example, \cite{lamport29} uses an $\ell_1$-norm relaxation commonly used for sparse vector estimation, yielding robustness to outliers \cite{decod,lamport25,lamport22}.
However, the approach presented in \cite{lamport29} has no provable guarantees and requires $\bC$ to be column sparse, i.e., a very small portion of the columns of $\bC$ can be non-zero. The work in \cite{lamport21} replaces the $\ell_1$-norm in \cite{lamport29} with the $\ell_{1,2}$-norm. While the algorithm in \cite{lamport21} can handle a large number of outliers, the complexity of each iteration is $\calO (n m^2)$ and its iterative solver has no performance guarantees.
 %The work in \cite{lamport21} modifies \cite{lamport29} by exploiting the column sparse structure of the outlier matrix (the column sparsity of $\bC$).
%However, these approaches are not directly applicable to large data matrices. %However, these methods are computationally expensive for large data matrices and there are no guarantees on their performance.
Recently, the idea of using a robust norm was revisited in \cite{hystack, zhang2014novel}. Therein, the non-convex constraint set is relaxed to a larger convex set and exact subspace recovery is guaranteed under certain conditions. The algorithm presented in \cite{hystack} obtains $\col(\bL)$ and \cite{zhang2014novel} finds its complement. % of the column space of $\bL$.
However, the iterative solver of \cite{hystack} computes a full  SVD of an $m \times m$ weighted covariance matrix in each iteration. Thus, the overall complexity of the solver of \cite{hystack} is roughly $\calO (m ^3  + n m^2)$ per iteration, where the second term is the complexity of computing the weighted covariance matrix. Similarly, the solver of \cite{zhang2014novel} has $\calO(n m^2 + m^3)$ complexity per iteration. % unlike most robust PCA algorithms which only need to compute a partial SVD.
In \cite{tsakiris2015dual}, the complement of the column space of $\bL$ is recovered via a series of linear optimization problems, each obtaining one direction in the complement space. This method is sensitive to structured outliers, particularly linearly dependent outliers, and requires the columns of $\bL$ not to exhibit a clustering structure, which prevails much of the real world data. % is not a practical assumption for real world data.
%the hard optimization problem (\ref{eq1}) is relaxed by replacing the non-convex constraint set with a larger convex set. Under certain conditions, it is shown that the exact subspace can be extracted.
Also, the approach presented in \cite{tsakiris2015dual} requires solving $m - r$ linear optimization problems consecutively resulting in high computational complexity and long run time for high-dimensional data.
%In addition, the assumption of \cite{zhang2014novel} is rather restrictive because it assumes that the columns of $\bD$ span the $N_1$-dimensional ambient space.

 Robust PCA using convex rank minimization  was first analyzed in \cite{lamport1,lamport22} for the element-wise corruption model.
In \cite{lamport10}, the algorithm analyzed in \cite{lamport1,lamport22} was extended to the column-wise corruption model where
it was shown that the optimal point of \begin{eqnarray}
\begin{aligned}
& \underset{\hat{\bL},\hat{\bC}}{\min}
& & \|\hat{\bL} \|_* + \lambda\|\hat{\bC}\|_{1,2} \\
& \text{subject to}
& & \hat{\bL} + \hat{\bC} = \bD \:
\end{aligned}
\label{eq6}
\end{eqnarray}
yields the exact subspace and correctly identifies the outliers provided that $\bC$ is sufficiently column-sparse. \textcolor{black}{The solver of (\ref{eq6})
requires too many iterations, each computing the SVD of an $m \times n$ dimensional matrix.} Also, the algorithm can only tolerate a small number of outliers -- the ratio $\frac{n_2}{n_1}$ should be roughly less than 0.05. Moreover, the algorithm is sensitive to linearly dependent outliers.
% it is a  weak algorithm in tolerating the presence of outlying columns ($\frac{n_2}{n_1}$ should be roughly less than 0.05).

A different approach to outlier detection was  proposed in \cite{soltanolkotabi2012geometric,lamport7}, where a data point is classified as an outlier if it does not admit a sparse representation in the rest of the data. However, this approach is limited to the randomly distributed unstructured outliers. In addition, the complexity of solving the corresponding optimization problem is $\calO(n^3)$ per iteration.
   %unlike the inliers which lie in a low dimensional subspace, a small number of them  are linearly dependent.
%This feature is leveraged effectively to distinguish the outlier data points in  \cite{lamport8,soltanolkotabi2012geometric}.
%While this approach can recover the correct subspace even if a remarkable portion of the data is outliers, it cannot detect linearly dependent outliers and has $\calO(n^3)$ complexity
% of its iterative solver is roughly
%per iteration \cite{lamport7}.
In the outlier detection algorithm presented in \cite{heckel2013robust}, a data point is identified as an outlier if the maximum value of its mutual coherences with the other data points falls below a predefined threshold.
Clearly, this approach places a restrictive assumption on the outlying data points and is unable to detect structured outliers. %  where they correspond  to interesting rare events in the data.

%outliers that lie in close neighborhoods. For instance, any repeated outliers will be falsely detected as inliers since their mutual coherence is 1.

\subsection{Motivation and summary of contributions}
This work is motivated by the limitations of prior work on robust PCA as summarized below. \\
% \begin{itemize}
% \item
\textcolor{black}{
\noindent\textbf{Complex iterations.} Most of the state-of-the-art robust PCA algorithms require a large number of iterations each with high computational complexity.  For instance, many of these algorithms require the computation of the SVD of an $m \times n$, or $m \times m$, or $n \times n$ matrix in each iteration \cite{lamport8,lamport10,hystack}, leading to long run time. % The iteration with high computation complexity along with large number of required iterations led to intolerable running time for high dimensional data.
}

% \item
\noindent\textbf{Guarantees.} While the optimal points of the optimization problems underlying many of the existing robust subspace recovery techniques yield the exact subspace, there are no such guarantees for their corresponding iterative solvers. Examples include the optimization problems presented in \cite{lamport10,lamport21}.
In addition, most of the existing guarantees are limited to the cases where the outliers are scattered uniformly at random in the ambient space and the inliers are distributed uniformly at random in $\col(\bL)$ \cite{soltanolkotabi2012geometric,tsakiris2015dual,heckel2013robust}.

% Although the performance of some of the robust subspace recovery algorithms are guaranteed to yield exact subspace, the exact recovery is not guaranteed with their iterative solver.
% For instance, the optimal points of the optimization problems presented in \cite{lamport10,lamport21} are proven to yield the exact subspace under some mild sufficient conditions, their iterative solvers (which are used in practice) are not supported with mathematical performance guarantee.
%For instance, although the optimal point of the optimization problems presented in \cite{lamport10,lamport21} is proven to yield exact subspace under some mild sufficient conditions, their iterative solvers (which are used in practice) are not supported with mathematical performance guarantee.
% \item

\noindent\textbf{Robustness issues.} % Some of the existing robust PCA algorithms are not based on a comprehensive definition for outlying data points.
Some of the existing algorithms are tailored to one specific class of outlier models. For example, algorithms based on sparse outlier models utilize sparsity promoting norms, thus can only handle a small number of outliers. On the other hand, algorithms such as \cite{heckel2013robust,tsakiris2015dual,soltanolkotabi2012geometric} can handle a large number of unstructured outliers, albeit they fail to locate structured ones (e.g., linearly dependent or clustered outliers).
%
% \item
Spherical PCA (SPCA) is  a non-iterative robust PCA  algorithm that is also scalable \cite{maronna2006robust}. In this algorithm, all the columns of $\bD$ are first projected onto the unit sphere $\mathbb{S}^{m - 1}$, then the subspace is identified as the span of the principal directions of the normalized data. However, in the presence of outliers, the recovered subspace is never equal to the true subspace and it significantly deviates from the underlying subspace when outliers abound.
% \end{itemize}

To the best of our knowledge, CoP is the first algorithm that addresses these concerns all at once. In the proposed method, we distinguish outliers from inliers by comparing their degree of coherence with the rest of the data. The advantages of the proposed algorithm are summarized below.
\begin{itemize}
\item CoP is a considerably simple non-iterative algorithm which roughly involves one matrix multiplication to compute the Gram matrix.

% \item CoP is robust to linearly dependent outliers.
\item CoP can tolerate a large number of unstructured outliers. It is shown that exact subspace recovery is guaranteed with high probability even if $\frac{n_1}{n_2}$ goes to zero provided that $ \frac{n_1}{n_2} \frac{m}{r} $ is sufficiently large.
\item CoP is robust to both structured and unstructured outliers with provable performance guarantees for both types of outlying data points.

\item CoP is notably and provably robust to the presence of additive noise.

\end{itemize}

\begin{algorithm}
\caption{CoP: Proposed Robust PCA Algorithm}
{
\textbf{Initialization}: Set $p = 1$ or $p = 2$.

\smallbreak
\textbf{1. Data Normalization:}
Define matrix $\bX \in \mathbb{R}^{m \times n}$ as $\bx_i = \bd_i/ \| \bd_i \|_2$.

\smallbreak
\textbf{2. Mutual Coherence Measurement}\\
\textbf{2.1} Define $\bG = \bX^T \bX$ and set its diagonal elements to zero.

\textbf{2.2} Define vector $\bp \in \mathbb{R}^{n}$, where $\bp(i) = \| \bg_i \|_p, i = 1, \ldots, n$.

\smallbreak
\textbf{3. Subspace Identification:} Construct matrix $\bY$ from the columns of $\bX$ corresponding to the largest elements of $\bp$ such that they span an ${r}$-dimensional subspace.

\smallbreak
\textbf{Output:} The columns of  ${\bY}$ are a basis for $\col(\bL)$.
}
\end{algorithm}

\section{Proposed Method}
In this section, we present the Coherence Pursuit algorithm and provide some insight into its characteristics. The main theoretical results are provided in Section \ref{sec:theor_results}. Algorithm 1 presents CoP along with the definitions of the used symbols. % Before we discuss the main results, we first provide some insight.
\begin{figure}[t!]
	\centering
    \includegraphics[width=0.45\textwidth]{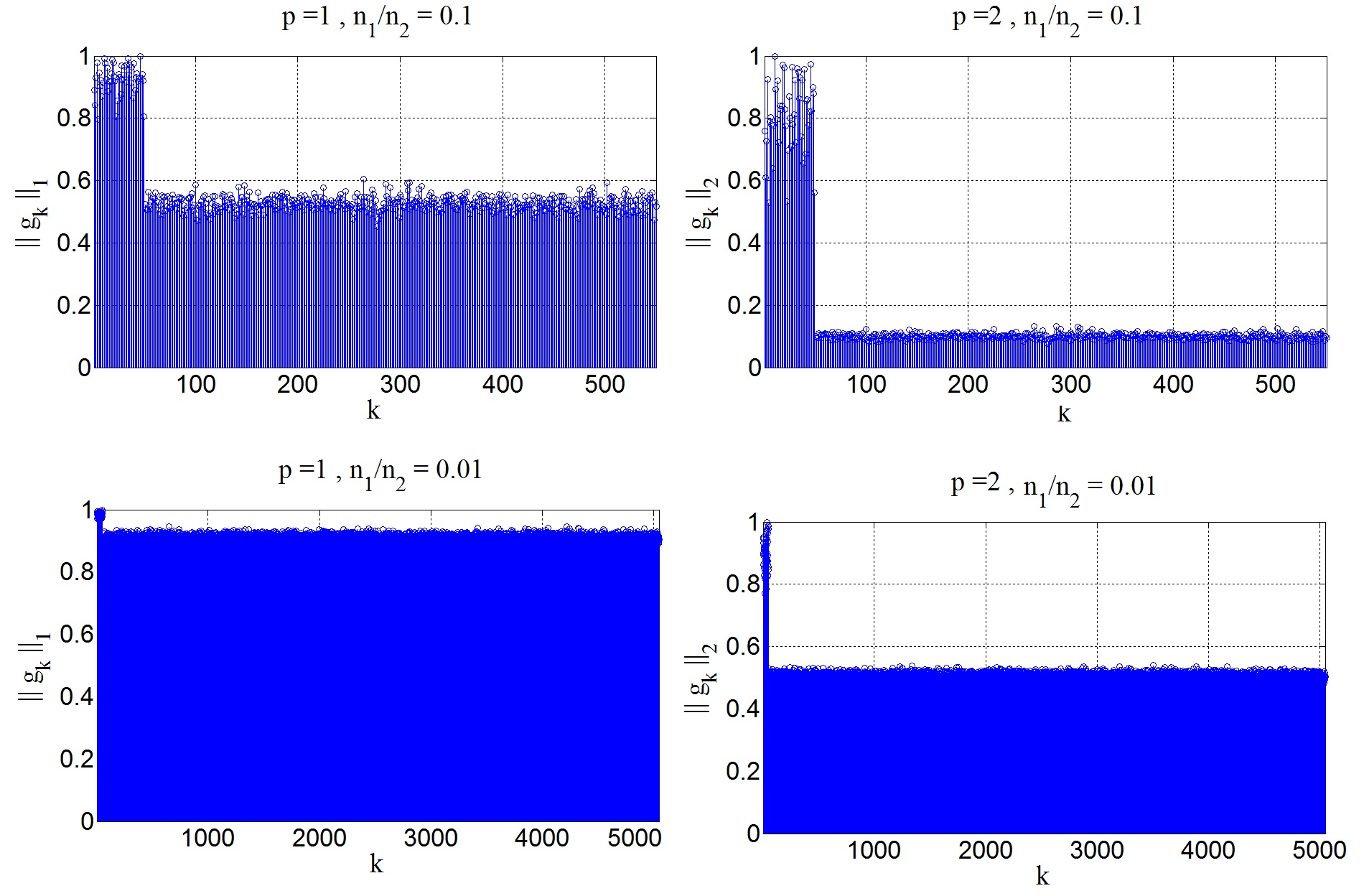}
    \vspace{-.1in}
    \caption{The values of vector $\bp$ for different values of $p$ and $\frac{n_1}{n_2}$. }
    \label{fig:differentp_n2n1}
\end{figure}
\smallbreak
\noindent{\it Coherence}: The inliers lie in a low-dimensional subspace $\calU$. In addition, the inliers are mostly correlated and form  clusters. Thus, an inlier bears strong resemblance to many other inliers. By contrast, an outlier is by definition dissimilar to most of the other data points.
As such, CoP uses the coherence value in Definition \ref{def:coh} to measure the degree of similarity between data points.
% how similar is a data point to the other data points.
Then, $\col(\bA)$ is obtained as the span of those columns that have large coherence values.

%Accordingly, the idea is to study the coherence of every data point with the other data points and obtain the column space of $\bA$ as the span of the columns which show strong coherence with the rest of the data.

For instance, assume that the distributions of the inliers and outliers follow the following assumption.

\begin{assumption}
The subspace $\calU$ is a random $r$-dimensional subspace in $\mathbb{R}^m$. % the $m$-dimensional ambient space.
The columns of $\bA$ are drawn  uniformly at random from the intersection
of $\mathbb{S}^{m - 1}$ and $\calU$. The columns of $\bB$ are drawn  uniformly at random from $\mathbb{S}^{m - 1}$. To simplify the exposition and notation, it is assumed without loss of generality that $\bT$ in (\ref{eq:data_model}) is the identity matrix, i.e, $\bD = [\bA \: \: \bB]$.
\label{assum_DistUni}
\end{assumption}
\noindent
Suppose the $i^{\text{th}}$ column is an inlier and the $(n_1 + j)^{\text{th}}$ column is an outlier.
In the appendix, it is shown that under Assumption \ref{assum_DistUni},  $\mathbb{E} \left[ \bp(i) \right] = \frac{n_1 - 1}{r} + \frac{n_2}{m}$, while $\mathbb{E} \left[ \bp(n_1 + j) \right] \leq \frac{r n1 + n_2}{m}$, for  $p=2$ where $\mathbb{E}[.]$ denotes the expectation.
Accordingly, if $m \gg r$, the inliers have much larger coherence values.
In the following, we demonstrate the important features of CoP, then present the theoretical results. % Subsequently, the theoretical results are presented.

\subsection{Large number of unstructured outliers}
Unlike some of the robust PCA algorithms which require $n_2$ to be much smaller than $n_1$, % or at least be smaller than $n_2$,
CoP tolerates a large number of unstructured outliers. For instance, consider a setting in which $m = 400$, $r = 5$, $n_1 = 50$, and the distributions of inliers and outliers follow Assumption \ref{assum_DistUni}.
Fig. \ref{fig:differentp_n2n1} shows the vector $\bp$ for different values of $p$ and $n_2$. In all the plots, the maximum element is scaled to 1. One can observe that even if $n_1/n_2 = 0.01$, CoP can  recover the exact subspace since there is a clear gap between the values of $\bp$ corresponding to outliers and inliers.

\begin{figure*}[t!]
	\centering
    \includegraphics[width=0.95\textwidth]{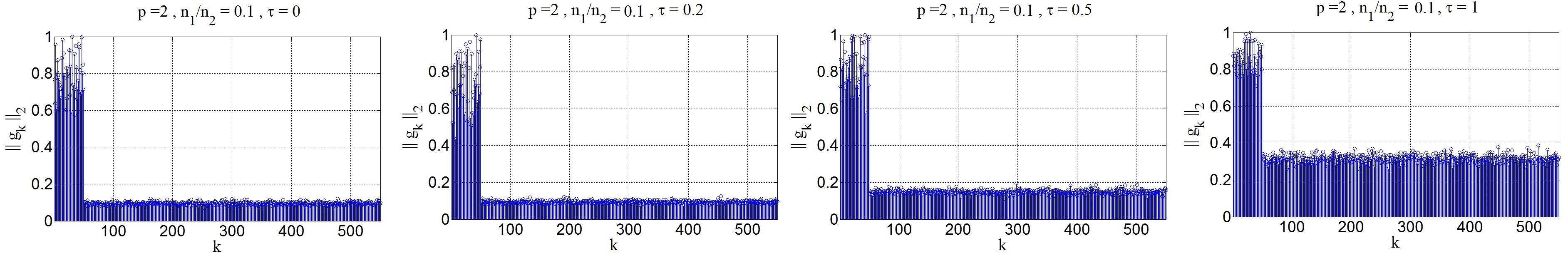}
    \vspace{-.1in}
    \caption{The elements of the vector $\bp$ for different values of the parameter $\tau$.}
    \label{fig:noise}
\end{figure*}

\subsection{Robustness to noise}
In the presence of additive noise, we model the data as
\begin{eqnarray}
\bD = [\bA \: \: \bB] \: \bT + \bE \:,
\end{eqnarray}
where  $\bE$ represents the noise
component.
%The columns of $\bA$ are  highly coherent together. However, the outliers, the columns of $\bB$,  are not expected to have strong resemblance with each other or with the inliers. Thus, there is a clear gap between the elements of $\bp$ corresponding to the inliers and the values corresponding to the outliers.

The strong resemblance between the inliers (columns of $\bA$) unlike the outliers (columns of $\bB$) creates a large gap between their corresponding coherence values as evident in Fig. \ref{fig:differentp_n2n1} even when $n_1/n_2 < 0.01$. This large gap affords tolerance to high levels of noise.
%
% For instance, one can observe from Fig. \ref{fig:differentp_n2n1} that even if $n_1/n_2 < 0.01$, there is a remarkable distance between the values corresponding to the inliers and the outliers. This gap helps the algorithm to be robust to additive noise because moderate amount of noise can not change the elements of $\bp$ remarkably.
% Our investigations show that the proposed algorithm is even robust to the strong presence of noise.
For example, assume $r=5$, $n_1 = 50$, $n_2 = 500$ and the distributions of the inliers and outliers follow Assumption \ref{assum_DistUni}. Define the parameter $\tau$ as
\begin{eqnarray}
\tau = \frac{\mathbb{E} \|\be \|_2 }{\mathbb{E} \|\ba \|_2}\:,
\end{eqnarray}
where $\ba$ and $\be$ are arbitrary columns of $\bA$ and $\bE$, respectively. Fig. \ref{fig:noise} shows the entries of $\bp$ for different values of $\tau$. As shown, the elements corresponding to inliers are clearly separated from the ones corresponding to outliers even at very low signal to noise ratio, e.g. $\tau = 0.5$ and $\tau=1$.

\subsection{\textcolor{black}{Structured outlying columns}}
%In many applications, the outliers could correspond to important rare events such as a cyber attack, malignant tissues, or important regions in an image. In these cases, the outlier could be linearly dependent or they could form data clusters with small population.
%Some of the existing robust PCA algorithms consider restrictive assumptions for the distribution of outliers and they cannot detect structured outliers \cite{tsakiris2015dual,heckel2013robust,soltanolkotabi2012geometric}.
%For instance, in the outlier detection algorithm presented in \cite{soltanolkotabi2012geometric} a data column is identified as an outlier if it does not admit a sparse representation in terms of the rest of the data. As such, if some outlying columns are repeated, the algorithm in \cite{soltanolkotabi2012geometric} fails to detect them.

At a fundamental level, CoP affords a global view of an outlying column, namely, a data column is identified as an outlier if it has weak total coherence with the rest of the data. This global view of a data point with respect to the rest of the data allows the algorithm to identify outliers that bear resemblance to few other outliers. Therefore, unlike some of the more recent robust PCA algorithms \cite{heckel2013robust,tsakiris2015dual,soltanolkotabi2012geometric} which are restricted to unstructured randomly distributed outliers, CoP can detect both structured and unstructured outliers. For instance, suppose the columns of $\bB$ admit the following clustering structure.
 \begin{assumption}
The $j^{\text{th}}$ outlier is formed as $\bb_i = \frac{1}{\sqrt{1+ \mu^2}} (\bq +\mu	 \bb_j^{'})$. The vectors $\bq$ and $\{ \bb_j^{'} \}_{j=1}^{n_2}$ are drawn uniformly at random from  $\mathbb{S}^{m-1}$.
\label{asm:outlier_clusters}
\end{assumption}

\begin{figure}[t!]
	\centering
    \includegraphics[width=0.5\textwidth]{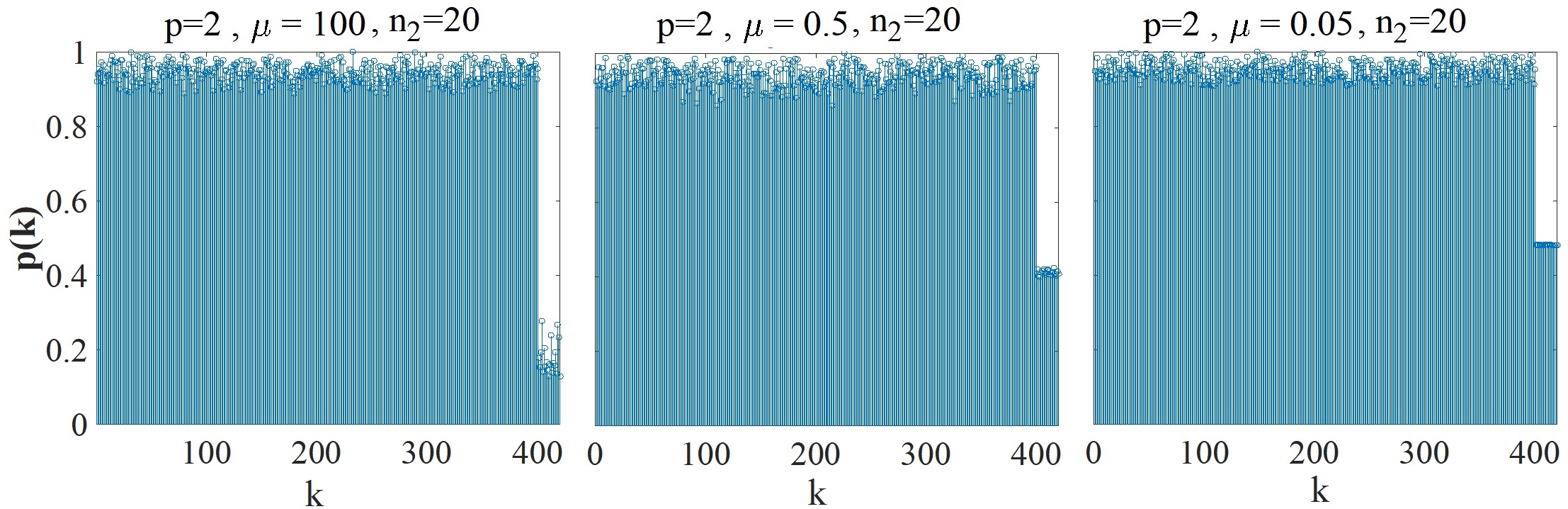}
    \vspace{-.2in}
    \caption{The data matrix $\bD \in \mathbb{R}^{200 \times 420}$ contains 20 outliers and the distribution of the outliers follows Assumption \ref{asm:outlier_clusters}. The elements of the vector $\bp$ are shown for different values of the parameter $\mu$. }
    \label{fig:dependent}
\end{figure}

\noindent
Under Assumption \ref{asm:outlier_clusters}, the columns of $\bB$ are clustered around $\bq$. As $\mu$ decreases, the outliers get closer to each other. Suppose $\bD \in \mathbb{R}^{200 \times 420}$ contains $20$ such outliers. % which follow Assumption \ref{asm:outlier_clusters}.
Fig. \ref{fig:dependent} shows the elements of $\bp$ for different values of $\mu$. When $\mu = 0.05$, the outliers are tightly concentrated around $\bq$, i.e., are very similar to each other, but even then CoP can clearly distinguish the outliers.
%The proposed method does not have a local view to detect an outlier. It considers the connection of a data point with all the rest of data. Accordingly, even if a data points is linearly dependent with few other data point, the proposed algorithm may identify it as an outlying data point.

%\textcolor{red}{
%For illustration, assume the given data follows Data model 1, $r = 5$, $n_1 = 50$, $n_2 = 500$, where the columns $301^{\text{th}}$ to $305^{\text{th}}$ are repeated outliers.
% Fig. \ref{fig:dependent} shows the elements of vector $\bp$ with $p=1$ and $p = 2$. All the elements corresponding to inliers are clearly greater than the elements corresponding to outliers and the algorithm correctly identifies the subspace.}
%\textcolor{black}{
%Fig. \ref{fig:dependent} suggests that CoP with $p = 1$ is better at handling repeated outliers since the entries of $\bG$ with large absolute magnitudes
%are more gracefully amplified by the $\ell_1$-norm.} %$\ell_1$-norm magnifies less the elements of $\bG$ with large absolute magnitude.}

\subsection{Subspace identification}
In the third step of Algorithm 1, we sample the columns of $\bX$ with the largest coherence values which span an $r$-dimensional space. In this section, we present several options for efficient implementation of this step.  One way is to start sampling the columns with the highest coherence values and stop when the rank of the sampled columns is equal to $r$. However, if the columns of $\bL$ admit a clustering structure and their distribution is highly non-uniform, this method will sample many redundant columns, which can in turn increase the run time of the algorithm. Hence, we propose two low-complexity techniques to accelerate the subspace identification step.

\noindent
1. In many applications, we may have an upper bound on $n_2/n$. For instance, suppose we know that up to 40 percent of the data could be outliers. In this case, we simply remove 40 percent of the columns corresponding to the smallest values of $\bp$ and obtain the subspace using the remaining data points.

\noindent
2. The second technique is an adaptive sampling method presented in Algorithm 2. First, the data is projected onto a random $ k r $-dimensional subspace to reduce the computational complexity for some integer $k > 1$. According to the analysis presented in \cite{rahmani2015randomized,lamport9}, even $k = 2$ is sufficient to preserve the rank of $\bA$ and the structure of the outliers $\bB$, i.e., the rank of $\mathbf{\Phi} \bA$ is equal to $r$ and the columns of $\mathbf{\Phi} \bB$ do not lie in $\col(\mathbf{\Phi} \bA)$, where $\mathbf{\Phi}$ is the projection matrix. The parameter $\upsilon$ that thresholds the $\ell_2$-norms of the columns of the projected data is chosen based on the noise level (if the data is noise free, $\upsilon = 0$).
In Algorithm 2, the data is projected onto the span of the sampled columns (step 2.3). Thus, a newly sampled column brings innovation relative to the previously sampled ones. Therefore, redundant columns are not sampled.

%In many application, we can roughly have an upper-bound on the number of outliers. For instance, if we know that $n_2/n$ is less than 0.1, we can reject 10 percent of the data corresponding to the smallest elements of $\bp$ and obtain the subspace and the span of the remaining data.

\begin{algorithm}
\caption{Adaptive Column Sampling for the Subspace Identification Step (step 3) of CoP}
{
\textbf{Initialization:} Set $k$ equal to an integer greater than 1, a threshold $\upsilon$ greater than or equal to $0$, and $\bF$ an empty matrix.

\smallbreak
\textbf{1. Data Projection:} Define $\bX_{\phi} \in \mathbb{R}^{k r \times n}$ as $\bX_{\phi} = \mathbf{\Phi} \bX$, where $ \mathbf{\Phi} \in \mathbb{R}^{k r \times m}$ projects the columns of $\bX$ onto a random $k r$-dimensional subspace.

\smallbreak
\textbf{2. Column Sampling}

\textbf{For} $i$ from 1 to $r$ do

\smallbreak
\textbf{2.1}
Define set $\calI$ as $\calI = \left\{ k \: \big | \: \| {\bX_{\phi} }_k \|_2 \leq \upsilon \right\}$. Set $\bp_{\calI} = 0$.
%Set equal to zero the elements of $\bp$ corresponding to columns of $\bX_{\phi}$ with $\ell_2$-norms less than or equal to $\nu$.

\textbf{2.2} Define $j := \underset{k}{\argmax}~\bp(k)$,  update $\bF = \text{orth} \Big( [\bF \:\: \bx_j] \Big)$, and set $\bp(j) = 0$.

\textbf{2.3} Update $\bX_{\phi} = \bX_{\phi} - \bF \bF^T \bX_{\phi}$.

\smallbreak

\textbf{End For}

\smallbreak

\textbf{Output} Construct $\bY$ using the columns of $\bX$ that correspond to the columns that formed $\bF$.
}
\end{algorithm}

\begin{remark}
Suppose we run Algorithm 2 ~$h$ times -- each time the sampled columns are removed from the data and newly sampled columns are added to $\bY$.  If the given data is noisy, the first $r$ singular values of $\bY$ are the dominant ones and the rest correspond to the noise component. If we increase $h$, the span of the dominant singular vectors will be closer to $\col(\bA)$. However, if $h$ is chosen unreasonably large, the sampler may also sample outliers.
\end{remark}

\subsection{Computational complexity}
The main computational complexity is in the second step of Algorithm 2 which is of order $\calO(m n^2)$. If we utilize Algorithm 2 as the third step of Algorithm 1, the overall complexity is of order $\calO(m n^2 + r^3 n)$.  However, unlike most existing algorithms, CoP does not require solving an optimization problem and roughly involves only one matrix multiplication. Therefore, it is very fast and simple for hardware implementation (c.f. Section \ref{sec: run time} on run time). Moreover, the overall complexity can be reduced to $\calO(r^4)$ if we utilize the randomized sketching designs presented in \cite{rahmani2015randomized,lamport9}.

\section{Theoretical Investigations}
\label{sec:theor_results}
% In order to find the important performance factors, we study the sufficient conditions which can guarantee the performance of the proposed method.
The theoretical results are presented in the next 4 subsections and their proofs are provided in Sections \ref{sec:proofs_main} and \ref{sec:proofs_sec}. First, we show that CoP can recover the true subspace even if the data is predominantly unstructured outliers. Second, we show that CoP can accurately distinguish structured outliers provided their population size is small enough. Third, we extend the robustness analysis to noisy settings. Fourth, we show that the more coherent the inliers are, the better CoP is at distinguishing them from the outliers.

In the theoretical studies corresponding to the unstructured outliers, the performance guarantees are provided for both $p=1$ and $p=2$. In the rest of the studies, the results are only presented for $p=1$. For each case, we present two guarantees.
First, we establish sufficient conditions to ensure that the expected values of the elements of the vector $\bp$ corresponding to inliers are much greater than the ones corresponding to outliers, in which case the algorithm is highly likely to yield exact subspace recovery. Second, we present theoretical results which guarantee exact subspace recovery with high probability.

\subsection{Subspace recovery with dominant unstructured outliers}
\label{sec:unstructured}
Here, we focus on the unstructured outliers, i.e., it is assumed that the distribution of the outlying columns follows Assumption \ref{assum_DistUni}.
The following lemmas establish sufficient conditions for the expected values of the elements of $\bp$ corresponding to inliers to be at least twice as large as those corresponding to outliers.
 % In the final version, we will refer to a technical report which contains the proofs}.
%The proofs of all theorems and lemmas are deferred to Sections \ref{sec:proofs_main} and \ref{sec:proofs_sec}.
%In order to show the important performance factors, we study the proposed method with the following random model for the distribution of the data which is a common random model in the literature of robust PCA [].

%According to our investigation, the proposed method is more robust to linearly dependent outliers with $p = 1$ but it can tolerate more amount of outliers with $p \ge 2$. In this section, we use $p = 1$ to analyze the algorithm. The following lemma shows that if $\frac{n_1}{\sqrt{r}}$ is sufficiently larger than $\frac{n_2 \sqrt{r}}{\sqrt{N_1}}$, the expected value of the $\ell_1$-norm of a column of $\bG$ corresponding to an inlier  is greater than the expected value of the $\ell_1$-norm of a column of $\bG$ corresponding to an outlier.

\begin{lemma}
Suppose Assumption 1 holds, the $i^{\text{th}}$ column is an inlier and the $(n_1 + j)^{\text{th}}$ column is an outlier. If
\begin{eqnarray}
\begin{aligned}
  \frac{n_1}{\sqrt{r}} \left( \sqrt{\frac{2}{\pi}} - \sqrt{\frac{4 r^2}{m}} \right)   > \frac{5 \: n_2}{4\sqrt{m}}  + \sqrt{\frac{2 }{\pi r}} \: ,
\end{aligned}
\label{suf:lemma1}
\end{eqnarray}
then $$ \mathbb{E} \: \|\bg_i \|_1 > 2 \: \mathbb{E} \: \|\bg_{n_1 + j} \|_1 \: $$ recalling that $\bg_i$ is the $i^{\text{th}}$ column of the Gram matrix $\bG$.
\label{lemma:expected}
\end{lemma}

\begin{lemma}
Suppose Assumption 1 holds, the $i^{\text{th}}$ column is an inlier and the $(n_1 + j)^{\text{th}}$ column is an outlier. If
\begin{eqnarray}
\begin{aligned}
  \frac{n_1}{r} ( 1 - \frac{2 r^2}{m} ) >   \frac{n_2 }{m} + \frac{1}{r}
\end{aligned}
\label{suff_lemm2}
\end{eqnarray}
then $$ \mathbb{E} \: \|\bg_i \|_2^2 > 2 \: \mathbb{E} \: \|\bg_{n_1 + j} \|_2^2 \: .$$
\label{lemma:expected 2}
\end{lemma}

%\begin{remark}
The sufficient conditions provided in Lemma \ref{lemma:expected} and Lemma \ref{lemma:expected 2} reveal three important points.
\smallbreak
\noindent I)
% Accordingly to the sufficient condition indictaed in Lemma \ref{lemma:expected 2} (and Lemma \ref{lemma:expected})
The ratios $\frac{n_1}{r}$ and  $\frac{n_2}{m}$ are key performance factors. The intuition is that
as $\frac{n_1}{r}$ increases, the density of the inliers in the subspace increases, and consequently their mutual coherence also increases. Similarly, the mutual coherence between the outliers is proportional to $\frac{n_2}{m}$.
% increases, so does the mutual coherence between the outliers.
 Thus, the main requirement is that $\frac{n_1}{r}$ should be sufficiently larger than $\frac{n_2}{m}$.
\smallbreak
\noindent II) In real applications, $r \ll m$ and $n_1 > n_2$, hence the sufficient conditions are easily satisfied. This fact is evident in Fig. \ref{fig:differentp_n2n1}, which shows that CoP can recover the correct subspace even if $n_1/n_2 = 0.01$.
\smallbreak
\noindent III) In high-dimensional settings, $r \ll m$. Therefore, $\frac{m}{\sqrt{m}}$ could be much greater than $\frac{r}{\sqrt{r}}$. Accordingly, the conditions in Lemma \ref{lemma:expected} are stronger than those in Lemma \ref{lemma:expected 2}, suggesting that CoP can tolerate a larger number of unstructured outliers with $p = 2$ than with $p =1$.
%Thus, it is expected that the proposed algorithm with $p = 2$ can tolerate more amount of outliers in comparison to the proposed method with $p =1$.
This is confirmed by comparing the plots in the last row of Fig. \ref{fig:differentp_n2n1}.
%\end{remark}

The following theorems show  that the same set of factors are important to guarantee that CoP recovers the exact subspace with high probability.

\begin{theorem}
If Assumption 1 is true and
\begin{eqnarray}
\begin{aligned}
\frac{n_1}{\sqrt{r}} &\left(\sqrt{\frac{2}{\pi}} - \frac{r + 2 \sqrt{\beta \kappa \: r}}{\sqrt{m}} \right) - 2\sqrt{n_1} - \sqrt{\frac{2 n_1 \log \frac{n_1}{\delta}}{r -1 }} \\
& >
 \frac{n_2}{\sqrt{m}} + 2\sqrt{n_2} + \sqrt{\frac{2 n_2 \log \frac{n_2}{\delta}}{m -1 }} + \frac{1}{\sqrt{r}}  \: ,
\end{aligned}
\label{sufficient_theorem3}
\end{eqnarray}
then Algorithm 1 with $p=1$ recovers the exact subspace with probability at least $1 - 3 \delta$, where $\beta = \max ( 8 \log  n_2/\delta , 8 \pi  ) $ and $\kappa = \frac{m}{m - 1}$.
\label{theorem1}
\end{theorem}

\begin{theorem}
If Assumption 1 is true and
\begin{eqnarray}
\begin{aligned}
& n_1 \left( \frac{1}{r} - \frac{r + 4  \zeta \kappa + 4 \sqrt{\zeta r \kappa} }{m} \right) - \eta_1  >
 2 \eta_2 + \frac{1}{r} \:,
\end{aligned}
\label{suff_theorem2}
\end{eqnarray}
then Algorithm 1 with $p=2$ recovers the correct subspace with probability at least $1-  4 \delta$,
where $$\eta_1 =  \max \left( \frac{4}{3} \log \frac{2r n_1 }{\delta} , \sqrt{4 \frac{n_1 }{r} \log \frac{2 r n_1}{\delta}} \right)\:,$$ $$\eta_2 =  \max \left( \frac{4}{3} \log \frac{2 m n_2}{\delta} , \sqrt{4 \frac{n_2}{m} \log \frac{2 m n_2}{\delta}} \right) \:, $$ $\zeta = \max (8 \pi , 8 \log \frac{n_2 }{\delta})$,  and $\kappa = \frac{m}{m - 1}$.
\label{theorem2}
\end{theorem}

\begin{remark}
The dominant factors of the LHS and the RHS of (\ref{suff_theorem2}) are $\frac{n_1}{r} \left( 1 - \frac{ r^2}{m} \right)$ and $\sqrt{4 \frac{n_2}{m} \log \frac{2 m n_2}{\delta}}$, respectively. As in Lemma \ref{lemma:expected 2}, we see the factor $\frac{n_2}{m}$, but under a square root. Thus, the requirement of Theorem \ref{theorem2} is less stringent than that of Lemma \ref{lemma:expected 2}.
This is because Theorem \ref{theorem2} guarantees that the elements of $\bp$ corresponding to inliers are greater than those corresponding to outliers with high probability, but does not guarantee a large gap between their values as in Lemma \ref{lemma:expected 2}.
% This is because the conditions of Lemma \ref{lemma:expected 2} ensure that the expected values of the elements of $\bp$ corresponding to the inliers are at least twice as large as the ones corresponding to the outliers. But, Theorem \ref{theorem2} guarantees that the elements corresponding to inliers are greater than those corresponding to outliers with high probability. It does not guarantee a notable gap between the values corresponding to the inliers and the outliers.
\end{remark}

%\section{Iterative Implementation}
%\bibliography{bibfile, refs}

%\bibliography{strings,refs}

\subsection{Distinguishing structured outliers}
\label{sec:structured}
In this section, we focus on structured outliers whose distribution is assumed to follow Assumption \ref{asm:outlier_clusters}.
%form a data cluster. The following assumption expresses the presumed model for the outlying data points.
Under Assumption \ref{asm:outlier_clusters}, each column has a unit expected squared norm, which affords a more tractable analysis versus normalizing the data.
%\begin{assumption}
%The $j^{\text{th}}$ outlier is formed as $\bb_i = \frac{1}{\sqrt{1+ \mu^2}} (\bq +\mu	 \bb_j^{'})$. The vectors $\bq$ and $\{ \ba_j^{'} \}_{j=1}^{n_2}$ are drawn uniformly at random from  $\mathbb{S}^{m-1}$.
%\label{asm:outlier_clusters}
%\end{assumption}
% According to Assumption \ref{asm:outlier_clusters},
The columns of $\bB$ are clustered around $\bq$, and get closer to each other as $\mu$ decreases.
% , the outliers become more close to each other.
The following lemma establishes that if $n_2$ is sufficiently small, the expected coherence value for an inlier is at least twice that of an outlier.

\begin{lemma}
Suppose the distribution of the outliers follows Assumption \ref{asm:outlier_clusters} and the inliers are distributed as in Assumption \ref{assum_DistUni}.
Define $\bG^{'} = \bD^T \bD$ and set the diagonal elements of $\bG^{'}$ equal to zero. Assume the $i^{\text{th}}$ column is an inlier, the $(n_1 + j)^{\text{th}}$ column is an outlier, and $\mu < 1$.
 If
\begin{eqnarray}
\begin{aligned}
& (n_1-1)\sqrt{\frac{2}{\pi r}} > \frac{2 n_2}{1+\mu^2} + \\
& \frac{1}{\sqrt{m}} \left( \frac{2\mu^2 n_2 + 4 \mu n_2 + 2n_1 \sqrt{r(1+\mu^2)} (\mu+1)}{1+\mu^2} \right) \:,
\end{aligned}
\label{eq:suff_coherent_data_out}
\end{eqnarray}
then $$ \mathbb{E} \:  \|\bg_i^{'} \|_1 > 2 \: \mathbb{E} \: \|\bg_{n_1 + j}^{'} \|_1 \: . $$

\label{lm:coherent_outliers}
\end{lemma}

\begin{figure}[t!]
	\centering
    \includegraphics[width=0.3\textwidth]{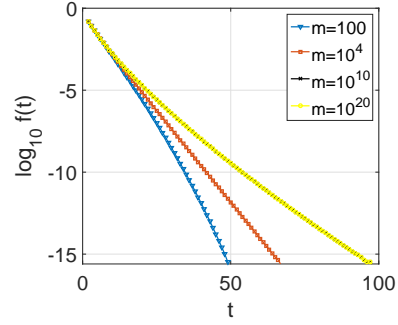}
    \vspace{-.06in}
    \caption{ The function $ \log_{10} f(t) = \log_{10} \left( 1 - \text{I}_{\frac{t}{m}} (0.5 , m/2 - 0.5) \right)$ versus $t$ for different values of $m$, where $\text{I}_{\frac{t}{m}} (0.5 , m/2 - 0.5)$ is the incomplete beta function.}
    \label{fig:incom_beta}
\end{figure}

\noindent
The sufficient condition (\ref{eq:suff_coherent_data_out}) is consistent with our intuition
regarding the detection of structured outliers. According to (\ref{eq:suff_coherent_data_out}), if $n_2$ is smaller or $\mu$ is larger (i.e., the outliers are less strcutured), the outliers will be more distinguishable.
The following theorem reaffirms the requirements of Lemma  \ref{lm:coherent_outliers}.
Before we state the theorem, we define $t_{\delta} := \inf\{t: f(t) < \delta\}$, where $f(t) = 1 - \text{I}_{\frac{t}{m}} (0.5 , m/2 - 0.5)$ and $\text{I}_{\frac{t}{m}} (0.5 , m/2 - 0.5)$ is the  incomplete beta function \cite{integrals_book}.
% let us define $t_{\delta}$ as if $t > t_{\delta}$, then $f(t) < \delta$ where $f(t) = 1 - \text{I}_{\frac{t}{m}} (0.5 , m/2 - 0.5)$. The function  $\text{I}_{\frac{t}{m}} (0.5 , m/2 - 0.5)$ is the  incomplete beta function \cite{integrals_book}.
The function $f(t)$ is monotonically decreasing.
% Roughly, the value of $f(t)$ exponentially decays with  $t$.
Examples are shown in Fig. \ref{fig:incom_beta}, which displays $\log_{10} f(t)$ for different values of $m$.
% of $\log_{10} f(t)$ For instance, Fig. \ref{fig:incom_beta} shows $\log_{10} f(t)$ for different values of $m$.
% It exponentially decays with $t$ and
The function $f(t)$ decays nearly exponentially fast with $t$ and converges for large values of $m$ to the function shown in yellow with circle markers in Fig. \ref{fig:incom_beta} where the plots for $m=10^{10}$ and $m=10^{20}$ coincide.

\begin{theorem}
Suppose the distribution of the outliers follows Assumption \ref{asm:outlier_clusters} and the inliers are as in Assumption \ref{assum_DistUni}.
Define $\bG^{'} = \bD^T \bD$ and set the diagonal elements of $\bG^{'}$ equal to zero. Assume the $i^{\text{th}}$ column is an inlier, the $(n_1 + j)^{\text{th}}$ column is an outlier, and $\mu < 1$.
 If
\begin{eqnarray}
\begin{aligned}
&  \sqrt{\frac{2}{\pi}} \frac{n_1 -1}{\sqrt{r}} - 2\sqrt{n_1} - \sqrt{\frac{2 n_1 \log \frac{n_1}{\delta}}{r }} >
 \frac{n_2}{1+ \mu^2 } + \\
 &\frac{\mu^2 + \mu}{1+\mu^2} \left( \frac{n_2 }{\sqrt{m}} + 2\sqrt{n_2} + \sqrt{\frac{2\: n_2  \log \frac{n_2}{\delta}}{m -1 }} \right) +  \\
 & \frac{\mu n_2 \sqrt{t_{\delta}}}{(1+\mu^2)\sqrt{m}}+\frac{n_1(\mu+1)}{\sqrt{(1+\mu^2)m}} \left( \sqrt{r} + 2 \sqrt{\beta \kappa} \right) \:,
\end{aligned}
\label{eq:suff_coherent_data2}
\end{eqnarray}
 then
$\| \bg_i^{'} \|_1 > \|\bg_{n_1 + j}^{'} \|_1 $ for all $ 1 \leq j \leq n_2$ and $ 1 \leq i \leq n_1$  with probability at least $1 - 6 \delta$, where $\beta = \max ( 8 \log  n_2/\delta , 8 \pi  ) $ and $\kappa = \frac{m}{m - 1}$.
\label{theorm:coherent_outliers}
\end{theorem}
% \noindent
Theorem \ref{theorm:coherent_outliers} certifies the requirements of Lemma \ref{lm:coherent_outliers}. According to (\ref{eq:suff_coherent_data2}), if the outliers are structured, the number of inliers should be sufficiently larger than the number of outliers.

\subsection{Performance analysis with noisy data}
\label{sec:noise_analysis}
CoP is notably robust to noise since the noise is neither coherent with the inliers nor the outliers. In this section, we establish performance guarantees for noisy data. It is assumed that the given data satisfies the following assumption.
\begin{assumption}
Matrices $\bA$ and $\bB$
 follow Assumption \ref{assum_DistUni}.
The columns of $\bE \in \mathbb{R}^{m \times n_1}$ are drawn  uniformly at random from $\mathbb{S}^{m - 1}$. Each column of matrix $\bA^e$ is defined as $\ba^e_i = \frac{1}{\sqrt{1+\sigma_n^2}} \left( \ba_i + \alpha_i \be_i \right)$, where $\{ \alpha_i \}_{i=1}^{n}$ are i.i.d samples from a normal distribution $\calN(0 , \sigma_n^2)$ and $\ba_i$ and $\be_i$ are the $i^{\text{th}}$ columns of $\bA$ and $\bE$, respectively. The given data matrix can be expressed as $\bD = [  \bA^e \: \: \bB].$
\label{assum_DistUni2}
\end{assumption}

\noindent
According to Assumption \ref{assum_DistUni2}, each inlier is a sum of a random unit $\ell_2$-norm vector in the subspace $\calU$ and a random vector $\alpha_i \be_i$ which models the noise.
% To make the problem analytically more tractable, in the following lemma and theorem, we do not normalize the data. However, in the model presumed by Assumption \ref{assum_DistUni2}, the expected value of the power of all the columns of data are equal to 1.
Per Assumption \ref{assum_DistUni2}, each data column has an expected squared norm equal to 1. % This assumption affords a more tractable analysis versus normalizing the data.
%In order to obtain concise results and simplify the analysis, we
%
\begin{lemma}
Suppose $\bD$ follows Asumption \ref{assum_DistUni2}. Define $\bG^e = \bD^T \bD$, set the diagonal elements of $\bG^e$ equal to zero, and define $\bp_e(k) = \| \bg^e_k \|_1$, where $\bg^e_k$ is the $k$-th column of $\bG^e$. In addition, assume the $i^{\text{th}}$ column is an inlier and the $(n_1 + j)^{\text{th}}$ column is an outlier.
If
\begin{eqnarray}
\begin{aligned}
&  \frac{n_1}{\sqrt{r}} \left( \sqrt{\frac{2}{\pi(1+\sigma_n^2)}} - \sqrt{\frac{4 r^2}{m}} \right)   > \\
  & \quad \quad\quad\quad\quad\quad\quad\quad\quad\quad \frac{ \: n_2 \sqrt{1+\sigma_n^2}  }{\sqrt{m}}  + \sqrt{\frac{2 }{\pi r}} + \xi \: ,
\end{aligned}
\label{eq:lm:suf_expect2}
\end{eqnarray}
where
\begin{eqnarray}
\xi \hspace{-.4mm}= \sqrt{\frac{2 \sigma_n^2}{\pi m}} \Bigg( \hspace{-1mm}\frac{n_1}{\sqrt{1+\sigma_n^2}} \left( 1 + \sigma_n \sqrt{\frac{\pi}{2}} + \sqrt{r} \right) + n_2 + 2n_1 \hspace{-1mm}\Bigg),
\end{eqnarray}
then $$ \mathbb{E} \: \|\bg_i^e \|_1 > 2 \: \mathbb{E} \: \|\bg_{n_1 + j}^e \|_1 \:. $$
\label{lm:Ewithnoise2}
\end{lemma}
The sufficient conditions of Lemma \ref{lm:Ewithnoise2} are very similar to the conditions presented in Lemma \ref{lemma:expected} for noise-free data with one main difference, namely, an additional term $\xi$ on the RHS of (\ref{eq:lm:suf_expect2}) due to the presence of noise. Nevertheless, akin to the unstructured outliers, the component corresponding to noise is linear in $1/\sqrt{m}$, where $m$ is the ambient dimension. In addition, $\sigma_n$ is practically smaller than 1 noting that the signal to noise ratio is $\frac{1}{\sigma_n^2}$. Thus, CoP exhibits robustness even in the presence of a strong noise component. The effect of noise is manifested in the subspace identification step wherein the subspace is recovered as the span of the principal singular vectors of the noisy inliers.
% presence of noise affects the subspace identification step which recovers the subspace as the span of principal singular vectors of the noisy inliers.
If the noise power increases, the distance between the span of the principal singular vectors of the noisy inliers and the column space of the noise-free inliers increases.  However, this error is inevitable and we cannot achieve better recovery given the noisy data. The following theorem affirms that the noise component does not have a notable effect on the sufficient conditions for the elements of $\bp_e$ corresponding to inliers to be greater than those corresponding to outliers with high probability.

\begin{theorem}
Suppose $\bD$ follows Asumption \ref{assum_DistUni2}. Define $\bG^e = \bD^T \bD$, set the diagonal elements of $\bG^e$ equal to zero, and define $\bp_e(k) = \| \bg^e_k \|_1$. If
\begin{eqnarray}
\begin{aligned}
\frac{n_1}{\sqrt{r}} &\left(\sqrt{\frac{2}{\pi (1+\sigma_n^2)}} - \frac{r + 2 \sqrt{\beta  \: r}}{\sqrt{m-1}} \right) \\
& - 2\sqrt{\frac{n_1}{1+\sigma_n^2}} - \sqrt{\frac{2 n_1 \log \frac{n_1}{\delta}}{(r -1)(1+\sigma_n^2) }} \\
& >
\sqrt{1+\sigma_n^2}\left( \frac{n_2}{\sqrt{m}} + 2\sqrt{n_2} + \sqrt{\frac{2 n_2 \log \frac{n_2}{\delta}}{m -1 }} \right) + \frac{1}{\sqrt{r}} +\varsigma \: ,
\end{aligned}
\label{eq:noise_suff}
\end{eqnarray}
where
\begin{eqnarray}
\begin{aligned}
& \varsigma = \left( \frac{c \sigma_n + c^2 \sigma_n^2}{\sqrt{1+\sigma_n^2}} + c\sigma_n \right) \left( \frac{n_1 }{\sqrt{m}} + 2\sqrt{n_1} + \sqrt{\frac{2\: n_1  \log \frac{n_1}{\delta}}{m -1 }} \right) \\
& + \frac{c\: n_1 \sigma_n}{\sqrt{1+\sigma_n^2}} \left( \sqrt{\frac{r}{m}} + 2\sqrt{\frac{\beta^{'}}{m-1}} \right) \:,
\end{aligned}
\end{eqnarray}
$c = \sqrt{2 \log \frac{n}{ \delta \sqrt{2 \pi} \sigma_n }}$, $\beta = \max(8\pi , 8 \log n_2 /\delta)$, and $\beta^{'} = \max(8\pi , 8 \log n_1 /\delta)$, then
$\| \bg_i^e \|_1 > \|\bg_{n_1 + j}^e \|_1 $ for all $ 1 \leq j \leq n_2$ and $ 1 \leq i \leq n_1$  with probability at least $1 - 8 \delta$.
\label{thm:withnoise}
\end{theorem}
Again, the sufficient condition (\ref{eq:noise_suff}) is very similar to (\ref{sufficient_theorem3}) for noise-free data. The main difference is the additional term $\varsigma$ on the RHS of (\ref{eq:noise_suff}). However, the presence of $\varsigma$ has no effect on the orders in the sufficient condition in comparison to (\ref{sufficient_theorem3}), and $\varsigma$ is approximately linear in $\sigma_n$.

%\begin{remark}
%Theorem \ref{theorem1} which provides the sufficient condition for noise free data roughly states  that $\frac{n_1}{\sqrt{r}}$ has to be sufficiently greater than $\frac{n_2}{\sqrt{m}} + \sqrt{n_2}$. According to Theorem \ref{thm:withnoise}, the requirement with noisy data changes to $\frac{n_1}{\sqrt{r}}$ has to be sufficiently greater than $ \left( \frac{n_2}{\sqrt{m}} + \sqrt{n_2} \right) (1 + \sigma_n)$. However, $\sigma_n$ is expected to be less than 1 since $\frac{1}{\sigma}$ is the signal to noise ratio.
%\end{remark}

\subsection{The distribution of inliers}
In the theoretical investigations presented in Section \ref{sec:unstructured}, Section \ref{sec:structured} , and Section \ref{sec:noise_analysis}, we assumed a random distribution for the inliers. However, we emphasize that this is not a requirement of the proposed approach.
% the proposed approach does not need the random distribution to perform properly.
In fact, the random distribution of the inliers leads to a fairly challenging scenario.
% creates a challenging scenario for CoP.
In practice, the inliers mostly form clusters and tend to be highly correlated. Since CoP exploits the coherence between the inliers, its ability to distinguish inliers could even improve if their distribution is further away from being uniformly random.
% is based on the coherency between the inliers and the less the data points are distributed randomly, the better CoP   distinguishes the inliers.
 We provide a theoretical example to underscore this fact. In this example, we assume that the inliers form a cluster around a given direction in $\calU$. The distribution of the inliers is formalized in the following assumption. % expresses the presumed distribution for inliers.

\begin{assumption}
The $i^{\text{th}}$ inlier is formed as $\ba_i = \frac{1}{\sqrt{1+ \nu^2}} (\bt +\nu	 \ba_i^{'})$. The vectors $\bt$ and $\{ \ba_i^{'} \}_{i=1}^{n_1}$ are drawn uniformly at random from the intersection of $\mathbb{S}^{m-1}$ and $\calU$.
\label{asm:inliers_clusters}
\end{assumption}

\begin{figure}[t!]
	\centering
    \includegraphics[width=0.5\textwidth]{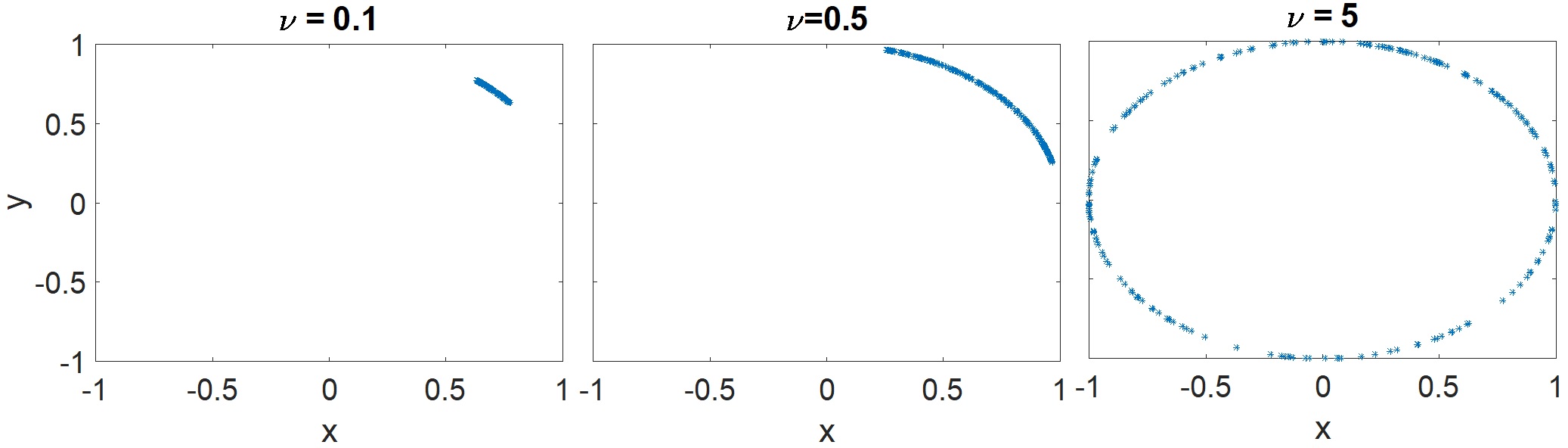}
    \vspace{-.02in}
    \caption{The distribution of inliers within $\calU$ for different values of parameter $\nu$ defined in Assumption \ref{asm:inliers_clusters}. If the value of $\nu$ increases, the inliers are less clustered. }
    \label{fig:inlier_clustered}
\end{figure}

\noindent
According to Assumption \ref{asm:inliers_clusters}, the inliers are clustered around the vector $\bt$. For example, suppose $r = 2$ and $n_1 = 200$. Fig. \ref{fig:inlier_clustered} shows the distribution of the inliers for different values of $\nu$. The data points become more uniformly distributed as $\nu$ increases, and from a cluster when $\nu$ is less than one.

\begin{lemma}
Suppose the distribution of the inliers follows Assumption \ref{asm:inliers_clusters} and the columns of $\bB$ are drawn uniformly at random from $\mathbb{S}^{m-1}$.
Define $\bG^{'} = \bD^T \bD$ and set its diagonal elements to zero. Assume the $i^{\text{th}}$ column is an inlier, the $(n_1 + j)^{\text{th}}$ column is an outlier, and $\nu < 1$.
 If
\begin{eqnarray}
\begin{aligned}
& n_1 \left( 1 - \frac{\nu^2 + 2 \nu}{\sqrt{r}}  \right) > 1+ \frac{2 \: n_1(1+\nu) \sqrt{r(1+\nu^2)}}{\sqrt{m}} \\
& + \frac{n_2 \sqrt{1+\nu^2}}{\sqrt{m}} \left(  \nu- \sqrt{\frac{2}{\pi}} + 2\sqrt{1+\nu^2} \right),
\end{aligned}
\label{eq:suff_coherent_data}
\end{eqnarray}
then $$ \mathbb{E} \:  \|\bg_i^{'} \|_1 > 2 \: \mathbb{E} \: \|\bg_{n_1 + j}^{'} \|_1 \: . $$

\label{lm:coherent_inliers}
\end{lemma}

\noindent
According to (\ref{eq:suff_coherent_data}), if $\nu$ decreases (i.e., the data points are less randomly distributed), it is more likely that CoP recovers the correct subspace. In other words, with CoP, clustered inliers are preferred over randomly distributed inliers.

% distr_nu

\begin{algorithm}
\caption{Subspace Clustering Error Correction Method}
{
\textbf{Input}: The matrices $\{ \hat{\bD}^i \}_{i = 1}^L$ are the clustered data (the output of a subspace clustering algorithm) and $L$ is the number of clusters.

\smallbreak
\textbf{Error Correction}\\
 \textbf{For} $k$ from 1 to $t$ do\\
\textbf{1} Apply the robust PCA algorithm to the matrices $\{  \hat{\bD}^i \}_{i = 1}^L$. Define the orthonormal matrices $\{  \hat{\bU}^i \}_{i = 1}^L$ as the learned bases for the inliers of $\{  \hat{\bD}^i \}_{i = 1}^L$, respectively.

\textbf{2} Update the data clustering with respect to the obtained bases $\{ \hat{\bU}^i \}_{i = 1}^L$ (the matrices $\{ \hat{\bD}^i \}_{i = 1}^L$ are updated), i.e., a data point $\bd$ is assigned to the $i^{\text{th}}$ cluster if $ i = \underset{k}{ \argmax} \: \| \bx^T \hat{\bU}^k \|_2$.
\\
\textbf{End For}

\smallbreak
\textbf{Output:} The matrices $\{ \hat{\bD}^i \}_{i = 1}^L$ represent the clustered data and the matrices $\{ \hat{\bU}^i \}_{i = 1}^L$ are the orthonormal bases for the identified subspaces.
}
\end{algorithm}

\section{Numerical Simulations}
In this section, the performance of CoP is investigated with both synthetic and real data. We compare the performance of CoP with the state-of-the-art robust PCA algorithms including FMS \cite{lerman2014fast}, GMS \cite{zhang2014novel}, R1-PCA \cite{lamport21}, OP \cite{lamport10}, and SPCA \cite{maronna2006robust}. \textcolor{black}{ For FMS, we implemented Algorithm 1 in \cite{lerman2014fast} with $p=0.5$. We have also tried different values for $p \leq 1$, which did not yield much difference in the results from what we report in our experiments.
For the GMS algorithm, we implemented Algorithm 2 in \cite{zhang2014novel} to obtain the matrix $\bQ$. The output of the algorithm is the last $r$ singular vectors of the obtained matrix $\bQ$, which serve as an orthonormal basis for the span of the inliers. For R1-PCA, we implemented the iterative algorithm presented in \cite{lamport21}, which iteratively updates an orthonormal basis for the inliers. }

\subsection{Phase transition}
\begin{figure}[t!]
	\centering
    \includegraphics[width=0.4\textwidth]{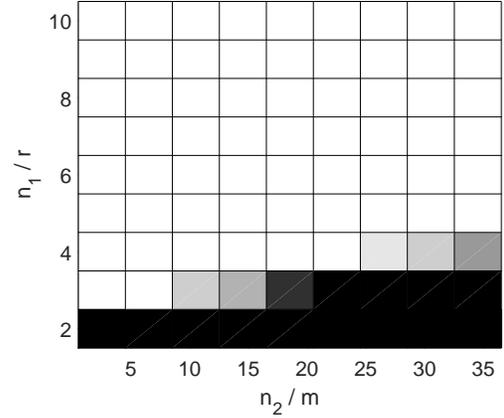}
    \vspace{-.08in}
    \caption{The phase transition plot of exact subspace recoveru in presence of unstructured outliers versus $n_1/r$ and $n_2/m$. }
    \label{fig:phase}
\end{figure}

Our analysis with unstructured outliers has shown that CoP yields exact subspace recovery with high probability if $n_1 /r$ is sufficiently greater than $n_2/m$. In this experiment, we investigate the phase transition of CoP in the $n_1 /r$ and $n_2/m$ plane. Suppose,
$m = 100$, $r = 10$, and the distributions of inliers/outliers follow Assumption 1.
Define $\bU$ and $\hat{\bU}$ as the exact and recovered orthonormal bases for the span of the inliers, respectively.
 A trial is considered successful if
$$ \left(  \| \bU  -  \hat{\bU} \hat{\bU}^T \bU \|_F / \| \bU \|_F \right) \leq 10^{-5} \:.$$
In this simulation, we construct the matrix $\bY$ using 20 columns of $\bX$ corresponding to the largest 20 elements of the vector $\bp$.
Fig. \ref{fig:phase} shows the phase transition, where white indicates correct subspace recovery and black designates incorrect recovery. As shown, if $n_2/m$ increases, we need larger values of $n_1/r$. However, one can observe that with $n_1/r > 4$, the algorithm can yield exact recovery even if $n_2/m > 30$.

\subsection{Running time }
\label{sec: run time}
In this section, we compare the speed of CoP with the existing approaches. Table \ref{tab:runnigtime} shows the run time in seconds for different data sizes. In all experiments, $n_1 = n/5$ and $n_2 = 4 n /5$. One can observe that CoP is remarkably faster by virtue of its simplicity (single step algorithm).

\begin{table}[h]
\centering
\caption{Running time of the algorithms}
\begin{tabular}{| c | c | c | c | c |}
\hline
$m = n$ & CoP &     FMS & OP & R1-PCA  \\
 \hline
  1000 & 0.02 & 1 & 45  & 1.45 \\
 \hline
 2000 & 0.7 & 5.6  & 133  &  10.3\\
    \hline
     5000 & 5.6 & 60  & 811  & 83.3\\
    \hline
     10000 & 27 & 401  & 3547  & 598 \\
    \hline
\end{tabular}
\label{tab:runnigtime}
\end{table}

\subsection{Subspace recovery in presence of unstructured outliers}
\begin{figure}[t!]
	\centering
    \includegraphics[width=0.45\textwidth]{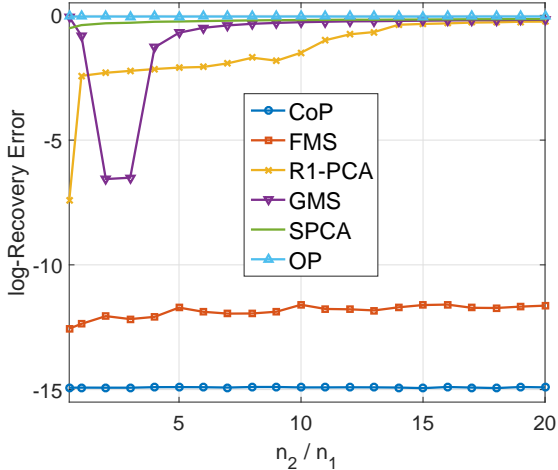}
    \vspace{-.02in}
    \caption{The subspace recovery error versus $n_2 / n_1$ for unstructured outliers. }
    \label{fig:recoveryerror}
\end{figure}

In this experiment, we assess the robustness of CoP to outliers in comparison to existing approaches. It is assumed that $m = 50$, $r = 10$, $n_1 = 50$ and the distribution of inliers/outliers follow Assumption 1. Define $\bU$ and $\hat{\bU}$ as before, % exact and obtained orthonormal bases for the span of inliers, respectively.
and the recovery error as
$$ \text{Log-Recovery Error} =  \log_{10} \left(  \| \bU  -  \hat{\bU} \hat{\bU}^T \bU \|_F / \| \bU \|_F \right) \:.$$
In this simulation, we use 30 columns to form the matrix $\bY$.
Fig. \ref{fig:recoveryerror} shows the recovery error versus $n_2 /n_1$ for different values of $n_2$. In addition to its simplicity, CoP  yields exact subspace recovery even if the data
is overwhelmingly outliers.
\textcolor{black}{
Similar to CoP and FMS, the algorithms presented in \cite{heckel2013robust,soltanolkotabi2012geometric}
can also yield exact subspace recovery in presence of unstructured outliers even if they dominate the data. However, they are not applicable to the next experiments that deal with structured outliers. For instance,  the outlier detection method presented in \cite{heckel2013robust} assumes  the order of the inner product between any two outliers is $\frac{\log n}{ \sqrt{m}}$. Therefore, it is unable to identify structured outliers in high-dimensional data.
}

\subsection{Detection of structured outliers}
In this section, we examine the ability of CoP at detecting structured outliers in four experiments. In the first experiment, a robust PCA algorithm is used to identify the €œsaliency map€ \cite{koch1987shifts} of a given image.  For the second experiment, an outlier detection algorithm is used to detect the frames corresponding to an activity in a video file. In the third, we
examine the performance of the robust PCA algorithms with synthetic structured outliers. For the fourth experiment, we consider the problem of identifying the dominant low-dimensional subspace with real world data.

\begin{figure}[t!]
	\centering
    \includegraphics[width=0.3\textwidth]{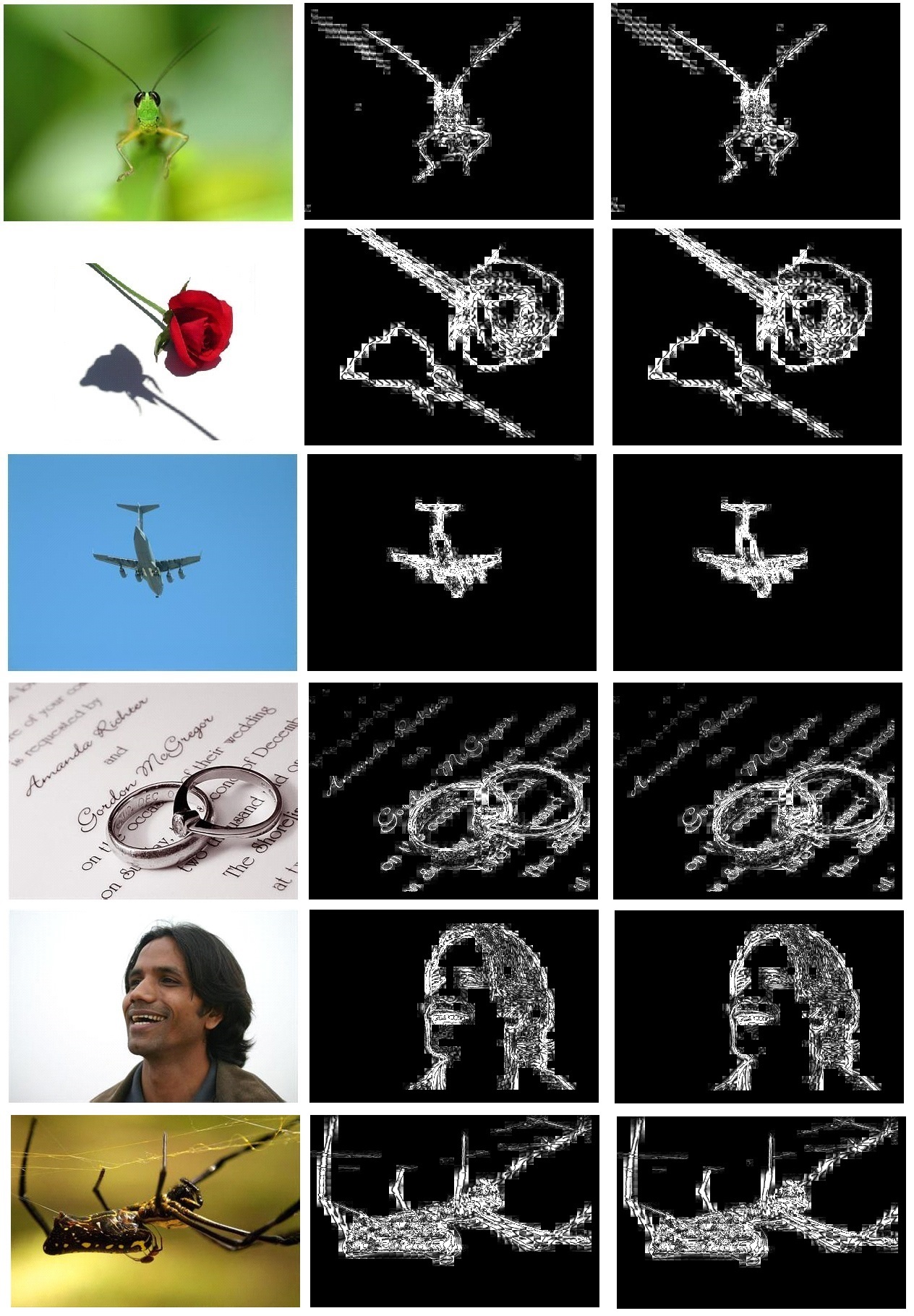}
    \vspace{-.02in}
    \caption{The first columns are the images obtained from the MSRA Salient Object Database. The second and third columns show the detection results obtained by CoP and FMS, respectively. }
    \label{fig:sail_map}
\end{figure}

\smallbreak
\noindent
\textbf{Example D.1 (Saliency map identification):} A saliency map indicates the regions of an image that tend
to attract the attention of a human viewer \cite{koch1987shifts,lamport9}. If we divide the image into small patches and construct a data matrix from the vectorized versions of the patches, the salient regions can be viewed as outlying columns \cite{lamport9,yu2011statistical1}. Hence, if we are able to detect the outlying columns, we will identify the salient regions from the corresponding patches. However, the different patches in the salient regions could be similar to each other. Therefore, the outlying data points are normally structured outliers. In this experiment, we obtained the images shown in the first column of Fig. \ref{fig:sail_map} from the MSRA Salient Object Database \cite{liu2011learning}.  The patches are non-overlapping $10\times10$ pixel windows. Fig. \ref{fig:sail_map} shows the saliency maps obtained by CoP and FMS. In both methods, the parameter $r$ (the rank of $\bL$) is set equal to 2.
As shown, both CoP and FMS properly identify the visually salient regions of the images since the two methods are robust to both structured and unstructured outliers.
%In this application, the outliers could be very similar to each other. For instance, Fig. () shows the elements of $\bh$ where $\bh(i) = \max \bg_i$. It shows that for eac

%The outlier detection algorithm presented in \cite{heckel2013robust} assumes that the outliers are distributted uniformly at random on the unit sphere. Thus, it expects that the inner product of an outlying column with any other data columns is small. According to our observations in this application, this is not an accurate assumption since  some of the patches corresponding to the salient regions are highly correlated. We set the threshold of \cite{heckel2013robust} for outlier detection equal to 0.98 as it appeared to yield the best performance. The only scenario where \cite{heckel2013robust} could be successful with clustered outliers is when there is a gap between the density of cluster of inliers and the cluster of outlier , the density of the cluster of outlier is less, and no two outlier are equal or very simialr.  In this case, if we can find an accurate value for the threshold value used in \cite{heckel2013robust}, we can distinguish the outliers. However, in practice we do not know this value a priori and  the value suggested in \cite{heckel2013robust} does not work with structured outliers.

 %\cite{heckel2013robust} can not identify some of the   salient regions. For instance, in the last image where we have repetitive patterns, the algorithm can not find the regions corresponding to the repeated patterns.

\begin{figure*}[t!]
	\centering
    \includegraphics[width=0.8\textwidth]{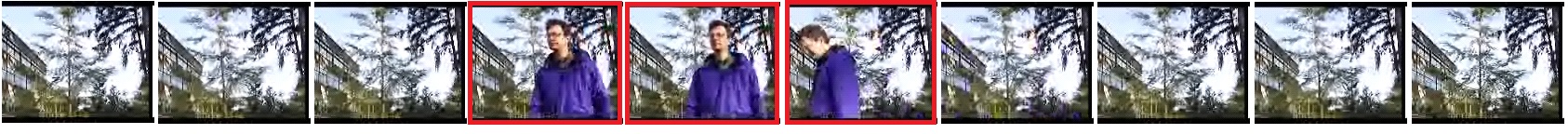}
    \vspace{-.02in}
    \caption{Some of the frames of the Waving Tree video file. The highlighted frames are detected as outliers by CoP and FMS.}
    \label{fig:activity}
\end{figure*}

\begin{figure}[t!]
	\centering
    \includegraphics[width=0.47\textwidth]{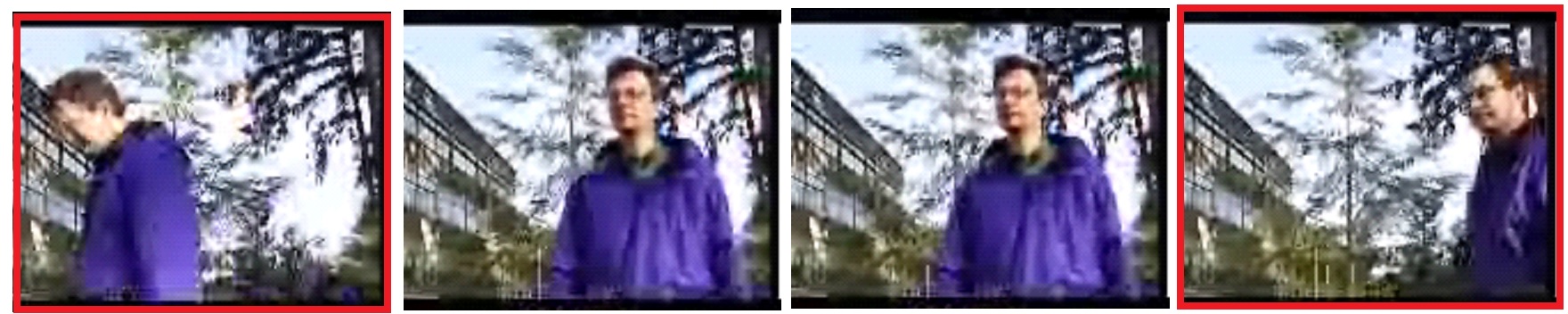}
    \vspace{-.02in}
    \caption{The highlighted frames indicate the frames detected as outliers by R1-PCA. }
    \label{fig:activity}
\end{figure}

\smallbreak
\noindent
\textbf{Example D.2 (Activity detection)}:  In many applications, an anomaly/outlier corresponds to the occurrence of some important rare event. In this experiment, we use the robust PCA method to detect activity in a video file. The file we use here is the Waving Tree file, a video of a dynamic background \cite{li2004statistical,toyama1999wallflower} showing a tree smoothly waving, and in the middle of the video a person crosses the frame. We expect the algorithm to detect those few frames where the person is present as outliers. We construct the data matrix from the vectorized video frames, i.e., each column corresponds to a specific frame.

The frames which show the background are inliers. Since the tree is waving, the rank of $\bL$ is greater than one. We set the parameter $r = 3$ in this experiment. The outliers correspond to the few frames in which the person crosses the scene. Obviously, in this application the outliers are structured because the consecutive frames are quite similar to each other. Thus, algorithms such as \cite{heckel2013robust,soltanolkotabi2012geometric}, which model the outliers as randomly distributed vectors, are not applicable here to detect the outliers. We use CoP, FMS, and R1-PCA to detect the outlying frames. Define $\hat{\bU} \in \mathbb{R}^{m \times 3}$ as the obtained orthonormal basis for the inliers.
We identify $\bd_i$ as an outlier if $ \| \bd_i - \hat{\bU} \hat{\bU} \bd_i \|_2 / \|\bd_i \|_2 > 0.2 $. CoP and FMS identify all the outlying frames correctly. Fig. \ref{fig:activity} shows some of the frames identified as inliers and outliers.
R1-PCA could only detect a subset of the outliers. In the video file, the person enters the scene from one side, stays for a second, and leaves the scene from the other side. R1-PCA detects only those frames in which the person enters or leaves the scene. Fig. \ref{fig:activity} shows two outlying frames that R1-PCA could detect and two frames it could not detect.

\noindent
\textbf{Example D.3 (Synthetic clustered outliers)}: In this experiment, we use synthetic data to study the performance of CoP in distinguishing structured outliers. The data matrix $\bD$ is generated as $\bD = [\bA \: \: \bB]$, where $\bA \in \mathbb{R}^{200 \times 400}$ with $r=5$ follows Assumption \ref{asm:inliers_clusters} with $\nu=0.2$. The matrix $\bB \in \mathbb{R}^{200 \times 20}$ follows Assumption \ref{asm:outlier_clusters}. Thus, the inliers
are clustered and the outliers could be clustered too depending on the value of $\mu$. Table \ref{tab:recovery_tab_error} shows the subspace recovery error, $ \| \bU  -  \hat{\bU} \hat{\bU}^T \bU \|_F / \| \bU \|_F$, for different values of $\mu$.
One can observe that CoP and GMS correctly recover the column space of $\bL$ for all values of $\mu$. However, for smaller values of $\mu$, where the outliers become more concentrated, FMS and R1-PCA fail to retrieve the exact subspace.

\smallbreak
\noindent
\textbf{Example D.4 (Dominant subspace identification):}
An application of robust PCA is in the problem of subspace clustering \cite{vidal2011subspaceff,rahmani2015innovation}. This problem is a general form of PCA in which the data points lie in a union of  linear subspaces \cite{vidal2011subspaceff}. A subspace clustering algorithm identifies the subspaces and clusters the data points with respect to the subspaces. A robust PCA algorithm can be applied in two different ways to the subspace clustering problem. The first way is to use the robust PCA method sequentially to learn one subspace in each iteration. In other words, in each iteration the data points in the dominant subspace (the one which contains the maximum number of data points) are considered as inliers and the others as outliers. In each step one subspace is identified and the corresponding data points are removed from the data.
RANSAC is a popular subspace clustering method which is based on robust PCA \cite{lamport48,vidal2011subspaceff}. The second way is to use robust PCA just to identify the most dominant subspace. In many applications, such as motion segmentation, the majority of the data points lie in a data cluster and the rest of the data points -- which are of particular interest -- form data clusters with smaller populations. Therefore, by identifying the dominant subspace and removing its data points, we can substantially reduce the computational complexity of the subsequent processing algorithms (e.g., the clustering algorithm).

In this experiment, we use the Hopkins155 dataset, which contains video sequences of 2 or 3 motions \cite{tron2007benchmark}. The data is generated by extracting and tracking a set of feature points through the frames. In motion segmentation, each motion
corresponds to one subspace. Thus, the problem here is to cluster data lying in two or three subspaces \cite{vidal2011subspaceff}.
Here, we use 8 data matrices of traffic videos with 2 motions. Since the data lies in a union of 2 subspaces, we can also cluster the data via learning the dominant subspace.
The number of data points in the dominant subspace is large and it is important to observe the accuracy of the algorithm at identifying the outliers. Thus, we define the average clustering error as
\[
\text{ACE} = 0.5 \left( n_1^{e}/n_1  + n_2^{e}/n_2  \right),
\]
where $n_1^{e}$ and $n_2^e$ are the numbers of misclassified inliers and misclassified outliers, respectively.
Table \ref{tab:motion_2_} reports the ACE for different algorithms. As shown, CoP yields the most accurate result. %It is worth noting that the outlier detection method presented in [] are not applicable to this

\begin{table}[h]
\centering
\caption{Subspace recovery error, $   \| \bU  -  \hat{\bU} \hat{\bU}^T \bU \|_F / \| \bU \|_F  $, of the algorithms versus the value of parameter $\mu$. }
\begin{tabular}{| c | c | c | c | c |}
\hline
$\mu$ & CoP &     FMS & GMS & R1-PCA  \\
 \hline
  5 & $< 10^{-5}$ & $< 10^{-5}$ & $< 10^{-5}$  & $< 10^{-5}$\\
 \hline
 0.5 & $< 10^{-5}$ & $< 10^{-5}$  & $< 10^{-5}$  &  0.15 \\
    \hline
     0.2 & $< 10^{-5}$ & 0.28  & $< 10^{-5}$  & 0.44\\
    \hline
     0.1 & $< 10^{-5}$ & 0.45 & $< 10^{-5}$  & 0.44 \\
    \hline
\end{tabular}
\label{tab:recovery_tab_error}
\end{table}

\begin{table}[h]
\centering
\caption{Average Clustering Error (ACE) of the algorithms for clustering the traffic data sequences with two motions (Mean - Median)}
\begin{tabular}{| c | c | c | c | c |}
\hline
 & CoP &     FMS & GMS & R1-PCA  \\
 \hline
  $ACE$ & 0.1 - 0.01 & 0.22 - 0.18 & 0.27 - 0.23  & 0.23 - 0.19\\
 \hline
\end{tabular}
\label{tab:motion_2_}
\end{table}

\subsection{Clustering error correction -- Real data}
%The subspace clustering problem is a general form of PCA in which the data points lie in a union of an unknown number of unknown linear subspaces \cite{vidal2011subspaceff}. A subspace clustering algorithm identifies the subspaces and clusters the data points with respect to the subspaces.
In this section, we present a new application of robust PCA in subspace clustering.
The performance of the subspace clustering algorithms -- especially the ones with scalable computational complexity -- degrades in presence of noise or when the subspaces are closer to each other. Without loss of generality,  suppose the data $ \bD = [\bD^1 \: ... \: \bD^L]$, where the columns of $\{ \bD^i \}_{i=1}^L$ lie in the linear subspaces $\{ \calS_i \}_{i=1}^L$, respectively, and $L$ is the number of subspaces. Define $\{ \hat{\bD}^i \}_{i = 1}^L$ as the output of the clustering algorithm (the clustered data). Define the clustering error as the ratio of misclassified points to the total number of data points.
With errors in clustering, some of the columns of $\hat{\bD}^i$ believed to lie in $\calS_i$ may actually belong to some other subspace. Such columns can be viewed as outliers in the matrix $\hat{\bD}^i$.  Accordingly, the robust PCA algorithm can be utilized to correct the clustering error.
We present Algorithm 3 as an error correction algorithm which can be applied to the output of any subspace clustering algorithm to reduce the clustering error. In each iteration, Algorithm 3 applies the robust PCA algorithm to the clustered data to obtain a set of bases for the subspaces. Subsequently, the obtained clustering is updated based on the obtained bases.

\begin{figure}[t!]
	\centering
    \includegraphics[width=0.4\textwidth]{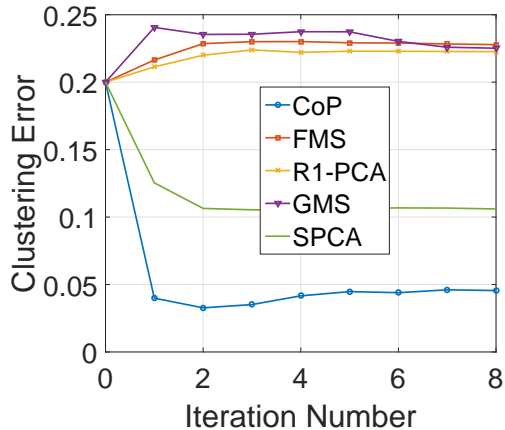}
    \vspace{-.02in}
    \caption{The clustering error after each iteration of Algorithm 2.}
    \label{fig:CE}
\end{figure}

In this experiment, we imagine a subspace clustering algorithm with 20 percent clustering error and apply Algorithm 3 to the output of the algorithm to correct the errors.
We use the Hopkins155  dataset.  Thus, the problem here is
to cluster data lying in two or three subspaces \cite{vidal2011subspaceff}. We use the traffic data sequences, which include 8 scenarios with two motions and 8 scenarios with three motions.
When CoP is applied, 50 percent of the columns of $\bX$ are used to form the matrix $\bY$.
Fig. \ref{fig:CE} shows the average clustering error (over all traffic data matrices) after each iteration of Algorithm 3 for  different robust PCA algorithms. CoP clearly outperforms the other approaches. As a matter of fact, most of the robust PCA algorithms fail to obtain the correct subspaces and end up increasing the clustering error.
The outliers in this application are linearly dependent and highly correlated. Thus, the approaches presented in \cite{heckel2013robust,tsakiris2015dual,soltanolkotabi2012geometric} which assume a random distribution for the outliers are not applicable.
%It worth noting that, in this application the outliers can be linearly dependent or they can be close together and the algorithm such as [] ans [] are not applicable to identify the

\section{Proofs of the Main Results}
\label{sec:proofs_main}
\noindent
\textbf{Proof of Lemma \ref{lemma:expected} }\\
The $i^{\text{th}}$ column of $\bG$ without its $i^{\text{th}}$ element can be expressed as
\begin{eqnarray}
[\ba_i^T \bA_{-i} \: \:\: \ba_i^T \bB]^T \:.
\end{eqnarray}
Thus,
\begin{eqnarray}
\| \bg_i \|_1 =  \| \ba_i^T \bA_{-i} \|_1 + \| \ba_i^T \bB \|_1 \:.
\label{eq:l1norm_gi}
\end{eqnarray}
If $i \neq k$, then
\begin{eqnarray}
\mathbb{E} | \ba_i^T \ba_k |  = \mathbb{E} | \bu^T \ba_k | \ge \sqrt{\frac{2}{\pi r}} \:,
\label{eq:exp1}
\end{eqnarray}
where $\bu$ is a fixed vector in $\calU$ with unit $\ell_2$-norm. The last inequality follows from \cite{hystack}. By Assumption 1, $\calU$ is a random subspace and $\ba_i$ is a random direction in $\calU$. Accordingly, the distribution of $\ba_i$ is the same as the distribution of a vector drawn uniformly at random from $\mathbb{S}^{m - 1}$.
Thus, similar to  (\ref{eq:exp1})
\begin{eqnarray}
\mathbb{E} | \ba_i^T \bb_k |  \ge
\sqrt{\frac{2}{\pi m}} \:.
\label{eq:exp2}
\end{eqnarray}
Replacing (\ref{eq:exp1}) and (\ref{eq:exp2}) in (\ref{eq:l1norm_gi}),
\begin{eqnarray}
\mathbb{E} \| \bg_i \|_1 \ge (n_1 -1) \sqrt{\frac{2}{\pi r}} + n_2 \sqrt{\frac{2}{\pi m}}.
\end{eqnarray}

The $(n_1 + j)^{\text{th}}$ column of $\bG$ without its $(n_1 + j)^{\text{th}}$ element can be expressed as
$$
[\bb_j^T \bA \: \:\: \bb_j^T \bB_{-j}]^T \:.
$$

\noindent
Define $\bU$ as an orthonormal basis for $\calU$. Thus,
\begin{eqnarray}
 \mathbb{E} | \bb_j^T \ba_k | \leq  \mathbb{E} \| \bb_j^T \bU \|_2.
\label{eq:proj}
\end{eqnarray}
It is not hard to show that
\begin{eqnarray}
 \mathbb{E} \| \bb_j^T \bU \|_2^2 = \frac{r}{m} \: .
\label{eq:expectequal}
\end{eqnarray}
Since $f(y)=y^2$ is a convex function, by Jensen's inequality
\begin{eqnarray}
 \mathbb{E} \| \bb_j^T \bU \|_2 \leq \sqrt{\frac{r}{m}} \: .
\label{eq:nim21}
\end{eqnarray}
Similarly, for $j \neq k$,
\begin{eqnarray}
 \mathbb{E} | \bb_j^T \bb_k | \leq \sqrt{\frac{1}{m}} \:.
\label{eq:nim22}
\end{eqnarray}
Therefore, according to (\ref{eq:nim21}) and (\ref{eq:nim22})
\begin{eqnarray}
 \mathbb{E} \| \bg_{n_1 + j} \|_1 \leq n_1 \sqrt{\frac{r}{m}} + (n_2 - 1) \sqrt{\frac{1}{m}}.
\end{eqnarray}
Thus, if (\ref{suf:lemma1}) is satisfied, $\mathbb{E} \| \bg_{i} \|_1 > 2 \mathbb{E} \| \bg_{n_1 + j} \|_1$.
\\
\\
\noindent
\textbf{Proof of Lemma \ref{lemma:expected 2}} \\
If the $i^{\text{th}}$ column is an inlier, then
\begin{eqnarray}
\| \bg_i \|_2^2 =  \| \ba_i^T \bA_{-i} \|_2^2 + \| \ba_i^T \bB \|_2^2 \:.
\label{eq:norm21}
\end{eqnarray}
Since the inliers are distributed uniformly at random within $\calU$,
\begin{eqnarray}
\mathbb{E} \| \ba_i^T \bA_{-i} \|_2^2 = \frac{n_1 -1}{r}.
\end{eqnarray}
The subspace $\calU$ is a random subspace and $\ba_i$ is a random direction within $\calU$. Thus,
\begin{eqnarray}
\mathbb{E} \| \ba_i^T \bB \|_2^2 = \frac{n_2}{m}.
\end{eqnarray}
Replacing in (\ref{eq:norm21}),
\begin{eqnarray}
\mathbb{E} \| \bg_i \|_2^2 = \frac{n_1 -1}{r} + \frac{n_2}{m} \:.
\end{eqnarray}
Similarly,
\begin{eqnarray}
\begin{aligned}
 \| \bg_{n_1 + j} \|_2^2 =  \| \bb_j^T \bA \|_2^2 + \| \bb_j^T \bB_{-j} \|_2^2.  \\
\end{aligned}
\label{eq:norm22}
\end{eqnarray}
Since $\calU$ is a random $r$-dimensional subspace,
\begin{eqnarray}
\mathbb{E} \: \| \bb_j^T \bA \|_2^2 \leq \frac{r \:  n_1}{m}
\end{eqnarray}
Accordingly,
\begin{eqnarray}
\begin{aligned}
& \mathbb{E} \: \| \bg_{n_1 + j} \|_2^2 \leq \frac{r \:  n_1}{m} + \frac{n_2 - 1}{m} \:.
\end{aligned}
\end{eqnarray}
Therefore, if (\ref{suff_lemm2}) is satisfied, $ \mathbb{E} \: \|\bg_i \|_2^2 > 2 \: \mathbb{E} \: \|\bg_{n_1 + j} \|_2^2 $.
\\
\\
\textbf{Proof of Theorem \ref{theorem1}}\\
We have to prove that the  $\ell_1$-norms of all the columns of $\bG$ corresponding to inliers are greater than the maximum of $\ell_1$-norms of the columns of $\bG$ corresponding to outliers. Thus, we establish a lower bound for $\{ \|\bg_k \|_1 \: \: | \: \:  1 \leq k \leq n_1 \}$
and an upper-bound for
$\{ \|\bg_k \|_1 \: \: | \: \:  n_1+1 \leq k \leq n_1+n_2 \}$ and it is shown that if (\ref{sufficient_theorem3}) is satisfied, the lower-bound is greater than the upper-bound with high probability. In order to establish the lower and upper bounds, we make use of the following lemmas.

\begin{lemma} \cite{hystack}
Suppose $\bg_1, ... ,\bg_{n}$   are i.i.d. random vectors distributed uniformly on the unit sphere $\mathbb{S}^{N - 1}$ in $\mathbb{R}^{N  }$. If $N > 2$, then
\begin{eqnarray}
\underset{\|\bu\| = 1}{\inf} \:\: \sum_{i = 1}^{n} | \bu^T \bg_i | > \sqrt{\frac{2}{\pi}} \frac{n}{\sqrt{N}} - 2\sqrt{n} - \sqrt{\frac{2 n \log \frac{1}{\delta}}{N -1 }}
\end{eqnarray}
with probability at least $1 - \delta$.
\label{lm:permenance}
\end{lemma}

\begin{lemma}
Suppose $\bg_1, ... ,\bg_{n}$   are i.i.d. random vectors distributed uniformly on the unit sphere $\mathbb{S}^{N - 1}$ in $\mathbb{R}^{N  }$. If $N > 2$, then
\begin{eqnarray}
\underset{\|\bu\| = 1}{\sup} \:\: \sum_{i = 1}^{n} | \bu^T \bg_i | <  \frac{n}{\sqrt{N}} + 2\sqrt{n} + \sqrt{\frac{2 n \log \frac{1}{\delta}}{N -1 }}
\label{suff_ng_perm}
\end{eqnarray}
with probability at least $1 - \delta$.
\label{lm:perm_neg}
\end{lemma}

\begin{lemma}  \cite{park2014greedy,ledoux2005concentration,milman2009asymptotic}
Let the columns of $\bF \in \mathbb{R}^{N \times {r}}$ be an orthonormal basis for an ${r}$-dimensional random subspace drawn uniformly at random in an ambient $N$-dimensional space. For any unit $\ell_2$-norm vector $\bc \in \mathbb{R}^{N \times 1}$
\begin{eqnarray}
\mathbb{P} \left[ \| \bc^T \bF \|_2 > \sqrt{\frac{{r}}{N}} + \sqrt{\frac{{8 \pi}}{N - 1}} + \sqrt{\frac{{8 \log 1/\delta}}{N - 1}} \right] \leq \delta \:.
\end{eqnarray}
\label{lm:projectranodm}
\end{lemma}

If the $i^{\text{th}}$ column of $\bX$ is an inlier, then $\| \bg_i \|_1 \ge \| \ba_i^T \bA_{-i} \|_1 $. Thus, the following corollary, which is based on Lemma \ref{lm:permenance}, establishes a lower-bound on the $\ell_1$-norm of a column of $\bG$ corresponding to an inlier.
\begin{corollary}
If Assumption 1 is true, then for all $ 1 \leq i \leq n_1$
\begin{eqnarray}
\| \ba_i^T \bA_{-i} \|_1 \ge \sqrt{\frac{2}{\pi}} \frac{n_1 -1}{\sqrt{r}} - 2\sqrt{n_1} - \sqrt{\frac{2 n_1 \log \frac{n_1}{\delta}}{r -1 }}
\label{suff_lema_int1}
\end{eqnarray}
with probability at least $1 - \delta$.
\label{lemma_inter1}
\end{corollary}

%%%%%%%%%%%%%%%%%%%%%%%%%%%%%%%
\noindent
\textbf{Proof of Corollary \ref{lemma_inter1}}
According to Lemma \ref{lm:permenance}
\begin{eqnarray}
\begin{aligned}
& \mathbb{P} \Bigg[ \| \ba_i^T \bA_{-i} \|_1 \ge \underset{\|\bu\| = 1}{\inf} \| \bu^T \bA_{-i} \|_1 \ge \\
&\sqrt{\frac{2}{\pi}} \frac{n_1 -1}{\sqrt{r}} - 2\sqrt{n_1} - \sqrt{\frac{2 n_1 \log \frac{n_1}{\delta}}{r -1 }} \Bigg]  \ge 1 - \delta/n_1 \:.
\end{aligned}
\label{inter_pro_c}
\end{eqnarray}
Thus, (\ref{suff_lema_int1}) is true for all $1 \leq i \leq n_1 $ with probability at least $1 - \delta$.

%%%%%%%%%%%%%%%%%%%%%%%%%%%%%%%

The $\ell_1$-norm of the $(n_1 + j)^{\text{th}}$ column of $\bG$ can be expressed as $ \left\| \: [\bb_j^T \bA \: \:\: \bb_j^T \bB_{-j}] \: \right\|_1$. Thus, the following two corollaries which are based on Lemma \ref{lm:perm_neg} and Lemma \ref{lm:projectranodm}  establish an upper-bound on $\| \bg_{n_1 + j} \|_1$.

\begin{corollary}
If Assumption 1 is true, then for all $ 1 \leq j \leq n_2$
\begin{eqnarray}
\| \bb_j^T \bB_{-j} \|_1 <  \frac{n_2 - 1}{\sqrt{m}} + 2\sqrt{n_2-1} + \sqrt{\frac{2\: n_2  \log \frac{n_2}{\delta}}{m -1 }}
\end{eqnarray}
with probability at least $1 - \delta$.
\label{lemma_inter2}
\end{corollary}

\begin{corollary}
If Assumption 1 is true, then for all $ 1 \leq j \leq n_2$
\begin{eqnarray}
\| \bb_j^T \bA \|_1 < n_1 \left(  \sqrt{\frac{r}{m}} + \sqrt{\frac{8 \pi}{m - 1}}  + \sqrt{\frac{8 \log  \frac{ n_2}{\delta} }{m -1 }}\right)
\label{suff_cor_por}
\end{eqnarray}
with probability at least $1 - \delta$.
\label{lemma_inter3}
\end{corollary}

%%%%%%%%%%%%%%%%%%%%%%%%%

\noindent
\textbf{Proof of Lemma \ref{lemma_inter3}}\\
The matrix $\bU$ is an orthonormal basis for $\col(\bA)$. Thus,
\begin{eqnarray}
\| \bb_j^T \bA \|_1 \leq n_1 \| \bb_j^T \bU \|_2 \: .
\end{eqnarray}
Thus, according to Lemma \ref{lm:projectranodm}, (\ref{suff_cor_por}) is true with probability at least $1 - \delta$.
%
%%%%%%%%%%%%%%%%%%%%%%%%%%%

According to Corollary \ref{lemma_inter1}, Corollary \ref{lemma_inter2} and Corollary \ref{lemma_inter3}, if (\ref{sufficient_theorem3}) is satisfied, then the proposed algorithm recovers the exact subspace with probability at least $1  - 3 \delta$.
\smallbreak
\noindent\textbf{Proof of Theorem \ref{theorem2}} \\
Theorem \ref{theorem2} can be proved using the results provided in the following lemmas and Lemma \ref{lm:projectranodm}.

\begin{lemma}
Suppose $\bg_1, ... ,\bg_{n}$   are i.i.d. random vectors distributed uniformly on the unit sphere $\mathbb{S}^{N - 1}$ in $\mathbb{R}^{N  }$. If $N > 2$, then

\begin{eqnarray}
\underset{\|\bu\| = 1}{\sup} \:\: \sum_{i = 1}^{n} (\bu^T \bg_i)^2 \leq \frac{n}{N} + \eta
\label{eq:suff_perm_2}
\end{eqnarray}
with probability at least $1 - \delta$ where  $ \eta = \max \left( \frac{4}{3} \log \frac{2N}{\delta} , \sqrt{4 \frac{n}{N} \log \frac{2 N}{\delta}} \right) $.
\label{lm:mylemma}
\end{lemma}

\begin{lemma}
Suppose $\bg_1, ... ,\bg_{n}$   are i.i.d. random vectors distributed uniformly on the unit sphere $\mathbb{S}^{N - 1}$ in $\mathbb{R}^{N  }$. If $N > 2$, then

\begin{eqnarray}
\underset{\|\bu\| = 1}{\inf} \:\: \sum_{i = 1}^{n} (\bu^T \bg_i)^2 > \frac{n}{N} -  \eta
\label{eq:suff_perm_2n}
\end{eqnarray}
with probability at least $1 - \delta$ where  $ \eta = \max \left( \frac{4}{3} \log \frac{2N}{\delta} , \sqrt{4 \frac{n}{N} \log \frac{2 N}{\delta}} \right) $.
\label{lm:mylemman}
\end{lemma}

\noindent
Based on Lemma \ref{lm:mylemma}, Lemma \ref{lm:mylemman} and Lemma \ref{lm:projectranodm}, we can establish the following corollaries from which Theorem \ref{theorem2} follows.

\begin{corollary}
If Assumption 1 holds, then for all $1 \leq i \leq n_1$
\begin{eqnarray}
 \mathbb{P} \left[ \| \ba_i^T \bA_{-i} \|_2^2 <
 \frac{n_1  - 1}{r} -  \eta \right] \leq \delta \:,
\end{eqnarray}
where  $ \eta = \max \left( \frac{4}{3} \log \frac{2r n_1}{\delta} , \sqrt{4 \frac{n_1 -1 }{r} \log \frac{2 r n_1}{\delta}} \right) $.
\label{lemma_inter12}
\end{corollary}

\begin{corollary}
If Assumption 1 holds, then for all $1 \leq i \leq n_1$
\begin{eqnarray}
 \mathbb{P} \left[ \| \ba_i^T \bB \|_2^2 <
 \frac{n_2}{m} -  \eta  \right] \leq \delta
\end{eqnarray}
where  $ \eta = \max \left( \frac{4}{3} \log \frac{2m }{\delta} , \sqrt{4 \frac{n_2}{m} \log \frac{2 m }{\delta}} \right) $.
\label{lemma_inter22}
\end{corollary}

\begin{corollary}
If Assumption 1 holds, then for all $1 \leq j \leq n_2$
\begin{eqnarray}
 \mathbb{P}
 \left[ \| \bb_j^T \bB_{-j} \|_2^2 >
 \frac{n_2 - 1}{m} + \eta \right] \leq \delta \:,
\end{eqnarray}
where  $ \eta = \max \left( \frac{4}{3} \log \frac{2m n_2}{\delta} , \sqrt{4 \frac{n_2 - 1}{m} \log \frac{2 m n_2 }{ \delta}} \right) $.
\label{lemma_inter32}
\end{corollary}

\begin{corollary}
If Assumption 1 holds, then for all $1 \leq j \leq n_2$
\begin{eqnarray}
\| \bb_j^T \bA \|_2^2\leq n_1 \left( \frac{r}{m} + \frac{4 \zeta}{m -1} + 4 \sqrt{\frac{\zeta r}{m (m -1 )}} \right)
\end{eqnarray}
with probability at least $1 - \delta$ where  $ \zeta = \max(8 \pi , \log \frac{n_2}{\delta}) $.
\label{lemma_inter32}
\end{corollary}

\smallbreak
\noindent
\textbf{Proof of Lemma \ref{lm:coherent_outliers}}\\
It suffices to ensure that
\begin{eqnarray}
\begin{aligned}
& \mathbb{E} \| \ba_i^T \bA_{-i} \|_1 > 2 \: \mathbb{E} \Big[ \frac{1}{\sqrt{1+\mu^2}} \| (\bq + \mu \bb_j^{'})^T \bA \|_1  \\
& +\frac{1}{{1+\mu^2}} \sum_{k=1 \atop k \neq j}^{n_2}  |(\bq + \mu \bb_j^{'})^T (\bq + \mu \bb_k^{'})| \Big] \: .
\end{aligned}
\label{eq:suff_out_coh}
\end{eqnarray}
Therefore, it is enough to guarantee that
\begin{eqnarray}
\begin{aligned}
& \mathbb{E} \| \ba_i^T \bA_{-i} \|_1 > 2 \: \mathbb{E} \Bigg[ \frac{1}{\sqrt{1+\mu^2}} \left( \|  \bq^T \bA \|_1 + \mu \|  {\bb_j^{'}}^T \bA \|_1 \right)  \\
& +  \frac{1}{{1+\mu^2}}\Bigg(  n_2 + \mu n_2 | \bq^T \bb_j^{'} | + \sum_{k=1 \atop k \neq j}^{n_2} \Big( \mu |\bq^T \bb_k^{'} |  \\
& \qquad\qquad\qquad+ \mu^2| {\bb_k^{'}}^T \bb_j^{'} | \Big)  \Bigg) \Bigg] \: .
\end{aligned}
\label{eq:suff_out_coh}
\end{eqnarray}
Similar to (\ref{eq:nim22}), $\mathbb{E} |\bq^T \bb_j^{'}| = \mathbb{E} |\bq^T \bb_k^{'}| =\mathbb{E} |{\bb_k^{'}}^T \bb_j^{'}| \leq 1/ \sqrt{m} $.
In addition, $\mathbb{E} \| \bq^T \bA \|_1 = \mathbb{E} \| {\bb_j^{'}}^T \bA \|_1 \leq n_1 \sqrt{r/m} $.
The rest of the proof is similar to the proof of Lemma \ref{lemma:expected}.

\smallbreak
\noindent
\textbf{Proof of Theorem \ref{theorm:coherent_outliers}}\\
The proof of Theorem \ref{theorm:coherent_outliers} is similar to the proof of Lemma \ref{lm:coherent_outliers}. We  use Corollary \ref{lemma_inter1} to lower bound the  LHS of (\ref{eq:suff_out_coh}). The random variable $|\bq^T \bb_j^{'}|^2$ follows the Beta distribution with parameter $\alpha = 1/2$, $\beta = m/2 - 1/2$, and $\mathbb{E} |\bq^T \bb_j^{'}|^2 = 1/m$ \cite{cai2012phase}.
According to the definition of $f(t)$,  $\mathbb{P} (|\bq^T \bb_j^{'}|^2 > \frac{t}{m})	= f(t)$. Thus, $\mathbb{P} (|\bq^T \bb_j^{'}| > \sqrt{\frac{t_{\delta}}{m})}	< \delta$. We also make use of Corollary \ref{lemma_inter2} and Corollary  \ref{lemma_inter3} to upper-bound the rest of the terms on the RHS of (\ref{eq:suff_out_coh}).

\smallbreak
\noindent
\textbf{Proof of Lemma \ref{lm:Ewithnoise2} }\\
The matrix $\bA^e$ can be expressed as $\bA^e = \frac{1}{\sqrt{1+\sigma_n^2}} (\bA + \bE^{'}) $ where $\be_i^{'} = \alpha_i \be_i$. If the $i^{\text{th}}$ column is an inlier, then
\begin{eqnarray}
\begin{aligned}
\| \bg_i^e \| &= \frac{1}{{1+\sigma_n^2}} \| (\ba_i + \alpha_i \be_i)^T(\bA_{-i} + \bE_{-i}^{'}) \|_1 \\
& +\frac{1}{\sqrt{1+\sigma_n^2}} \| (\ba_i + \alpha_i \be_i)^T \bB \|_1 \:.
\end{aligned}
\label{eq:exptnoise 1}
\end{eqnarray}
The first component of (\ref{eq:exptnoise 1}) can be lower bounded as
\begin{eqnarray}
\begin{aligned}
& \| (\ba_i + \alpha_i \be_i)^T(\bA_{-i} + \bE_{-i}^{'}) \|_1 \ge\\
& \|\ba_i \bA_{-i} \|_1 - \| \ba_i^T \bE_{-i}^{'} \|_1  - \| \alpha_i \be_i^T \bE_{-i}^{'} \|_1 - \| \alpha_i \be_i^T \bA_{-i} \|_1.
\end{aligned}
\label{eq:expanexp1}
\end{eqnarray}
According to (\ref{eq:expanexp1}) and similar to (\ref{eq:exp2}) and  (\ref{eq:nim22}),
\begin{eqnarray}
\begin{aligned}
& \mathbb{E} \left[ \Big\| (\ba_i + \alpha_i \be_i)^T(\bA_{-i} + \bE_{-i}^{'}) \Big\|_1 \right] \ge \\
& \left( \sqrt{\frac{2}{\pi r}}  - \sigma_n \sqrt{\frac{2}{\pi m}} - \sigma_n^2 \sqrt{\frac{1}{m}} - \sigma_n \sqrt{\frac{2 r}{m \pi}}   \right) (n_1 - 1) \:.
\end{aligned}
\label{eq:expanexp12}
\end{eqnarray}
Similarly,
\begin{eqnarray}
\begin{aligned}
\mathbb{E} \left[\| (\ba_i + \alpha_i \be_i)^T \bB \|_1\right] \ge \left( \sqrt{\frac{2}{\pi m}} - \sigma_n \sqrt{\frac{2}{m \pi}} \right) n_2 \:.
\end{aligned}
\end{eqnarray}

Similar to (\ref{eq:norm22}),
\begin{eqnarray}
\begin{aligned}
&  \| \bg_{n_1 + j} \|_1 =  \frac{1}{\sqrt{1+\sigma_n^2}} \| \bb_j^T (\bA + \bE^{'}) \|_1 + \| \bb_j^T \bB_{-j} \|_1 \\
 & \leq \frac{1}{\sqrt{1+\sigma_n^2}} \left( \|\bb_j^T \bA\|_1 + \| \bb_j^T \bE^{'} \| \right) + \| \bb_j^T \bB_{-j} \|_1 \:.
\end{aligned}
\label{eq:right2}
\end{eqnarray}
Thus, the expected value of (\ref{eq:right2}) can be upper bounded as
\begin{eqnarray}
\begin{aligned}
& \mathbb{E}  \| \bg_{n_1 + j} \|_1  \\
& \leq  \sqrt{\frac{1 }{m (1+\sigma_n^2)}} \left( n_1 \sqrt{r} +  n_2 \sqrt{1+\sigma_n^2}+ \sigma_n \sqrt{\frac{2}{\pi}} n_1  \right).
\end{aligned}
\label{eq:right22}
\end{eqnarray}
Thus, if (\ref{eq:lm:suf_expect2}) is satisfied, $\mathbb{E} \| \bg_{i}^e \|_1 > 2 \mathbb{E} \| \bg_{n_1 + j}^e \|_1$.

\smallbreak
\noindent
\textbf{Proof of Theorem \ref{thm:withnoise}}\\
It is sufficient to show that
\begin{eqnarray}
\begin{aligned}
& \frac{1}{{1+\sigma_n^2}} \Big( \| \ba_i^T \bA_{-i}\|_1 - \| \ba_i^T \bE^{'}_{-i} \|_1 \\
& \qquad\qquad\qquad - \| \alpha_i \be_i^T \bA_{-i}\|_1 - \| \alpha_i \be_i^T \bE_{-i}^{'}\|_1 \Big) \\
& > \frac{1}{\sqrt{1+\sigma_n^2}} \left( \| \bb_j^T \bA \|_1 + \| \bb_j^T \bE^{'} \|_1 \right) + \| \bb_j^T \bB_{-i} \|_1  \:.
\end{aligned}
\end{eqnarray}

\noindent
Define $\omega := \sigma_n \sqrt{2 \log \frac{n}{ \delta \sqrt{2 \pi} \sigma_n }}$. Thus, $\underset{i}{\max} \: \{ | \alpha_i | \}_{i=1}^n \leq \omega$ with probability at least $\delta$.
In order to obtain  the sufficient conditions, we make use of Corollary \ref{lemma_inter1}, Corollary \ref{lemma_inter2}, Corollary \ref{lemma_inter3}, and Corollary \ref{col:noise} stated below. Corollary \ref{col:noise} is derived using Lemma \ref{lm:perm_neg} and Lemma \ref{lm:projectranodm}.
\begin{corollary}
If Assumption \ref{assum_DistUni2}
 is true, then for all $ 1 \leq j \leq n_2$ and $ 1 \leq i \leq n_1$
\begin{eqnarray}
\begin{aligned}
& \frac{1}{\omega}  \| \ba_i^T \bE^{'}_{-i} \|_1 \leq \frac{n_1 - 1}{\sqrt{m}} + 2\sqrt{n_1-1} + \sqrt{\frac{2\: n_1  \log \frac{n_1}{\delta}}{m -1 }} \: ,
\\
&\frac{1}{\omega^2}  \| \alpha_i \be_i^T \bE^{'}_{-i} \|_1 \leq \frac{n_1 - 1}{\sqrt{m}} + 2\sqrt{n_1-1} + \sqrt{\frac{2\: n_1  \log \frac{n_1}{\delta}}{m -1 }} \: , \\
&  \frac{1}{\omega} \| \alpha_i \be_i^T \bA_{-i} \|_1 \leq (n_1 -1) \left(  \sqrt{\frac{r}{m}} + 2\sqrt{\frac{\beta^{'}}{m - 1}}  \right) \\
&  \frac{1}{\omega}  \| \bb_j^T \bE^{'} \|_1 \leq   \frac{n_1 }{\sqrt{m}} + 2\sqrt{n_1} + \sqrt{\frac{2 n_1 \log 1/\delta}{m -1 }} \: ,
\end{aligned}
\end{eqnarray}
 with probability at least $1- 5 \delta$, where $\beta^{'} = \max(8\pi , 8 \log n_1 /\delta)$ and $\omega = \sigma_n \sqrt{2 \log \frac{n}{ \delta \sqrt{2 \pi} \sigma_n }}$.
 \label{col:noise}
\end{corollary}

\smallbreak
\noindent
\textbf{Proof of Lemma \ref{lm:coherent_inliers}}\\
According to Assumption \ref{asm:inliers_clusters},
\begin{eqnarray}
\begin{aligned}
 & \|\bg_i^{'} \|_1 \ge \frac{1}{{1+\nu^2}} \sum_{k=1 \atop k \neq i}^{n_1} \Big[ |\bt^T \bt| - \nu^2 | {\ba_i^{'}}^T \ba_k^{'} | - \nu | \bt^T {\ba_i^{'}} |  - \\
 & \nu | \bt^T \ba_k^{'} | \Big]
 + \frac{1}{\sqrt{1+\nu^2}} \sum_{k=1}^{n_2} \left[ | \bt^T \bb_k | -  \nu|\bb_k^T\ba_i^{'}  | \right] \:.
\end{aligned}
\end{eqnarray}
In addition,
\begin{eqnarray}
\begin{aligned}
 & \|\bg_{n_1 + j}^{'} \|_1 = \frac{1}{\sqrt{1 + \nu^2}} \sum_{k=1}^{n_1}  \left|\bb_j^T (\bt + \nu \ba_k^{'}) \right| + \sum_{k=1 \atop k \neq j}^{n_2} | \bb_j^T \bb_k | \\
 & \leq \frac{\nu+1}{\sqrt{1 + \nu^2}} \sum_{k=1}^{n_1}  \| \bb_j^T \bU \| + \sum_{k=1 \atop k \neq j}^{n_2} | \bb_j^T \bb_k | \:.
\end{aligned}
\end{eqnarray}
The vectors $\bt$ and $\{ \ba_k^{'} \}_{k=1}^{n_1}$ are random vectors lying in $\calU$. Thus, $\mathbb{E} |{\ba_i^{'}}^T \ba_k^{'}| = \mathbb{E} |\bt^T \ba_i^{'}| \leq 1/\sqrt{r}$.
% Therefore, according to (\ref{eq:exp1}) and (\ref{eq:nim21}), if (\ref{eq:suff_coherent_data}).
In addition, $\mathbb{E} \| \bb_j^T \bU \|_2 \leq \sqrt{\frac{r}{m}}$ and $\mathbb{E} |\bb_j^T \bb_k| \leq \frac{1}{\sqrt{m}}$.
Thus, if (\ref{eq:suff_coherent_data}) is satisfied, then $ \mathbb{E} \:  \|\bg_i^{'} \|_1 > 2 \: \mathbb{E} \: \|\bg_{n_1 + j}^{'} \|_1 \:  $.

\section{Proofs of Intermediate Results}
\label{sec:proofs_sec}
\smallbreak
\smallbreak

\noindent
\textbf{Proof of Lemma \ref{lm:perm_neg}}\\
The proof of this Lemma is similar to the proof of Lemma \ref{lm:permenance} provided in \cite{hystack}. First, we add and subtract the mean values to expand (\ref{suff_ng_perm}) as follows
\begin{eqnarray}
\begin{aligned}
& \underset{\|\bu\| = 1}{\sup} \:\: \sum_{i = 1}^{n} | \bu^T \bg_i | \leq\\
& \underset{\|\bu\| = 1}{\sup} \:\: \sum_{i = 1}^{n} \left[ | \bu^T \bg_i | - \mathbb{E} | \bu^T \bg_i | \right] + \underset{\|\bu\| = 1}{\sup} \:\: \sum_{i = 1}^{n} \mathbb{E} | \bu^T \bg_i |
\end{aligned}
\label{eq:prv1}
\end{eqnarray}
Similar to (\ref{eq:nim22}),
\begin{eqnarray}
\underset{\|\bu\| = 1}{\sup} \:\: \sum_{i = 1}^{n} \mathbb{E} | \bu^T \bg_i | \leq n \sqrt{\frac{1}{N}} \:.
\label{eq:prv2}
\end{eqnarray}
Now, if we take similar steps used to prove Lemma B.3 in \cite{hystack}, the first component of (\ref{eq:prv1}) can be bounded as
\begin{eqnarray}
\begin{aligned}
 & \mathbb{P} \Bigg[ \underset{\|\bu\| = 1}{\sup} \:\: \sum_{i = 1}^{n} \left[ | \bu^T \bg_i | - \mathbb{E} | \bu^T \bg_i | \right] \ge 2 \sqrt{n} \\
 & \quad \quad \quad \quad\quad\quad + t \sqrt{\frac{n}{N - 1}} \Bigg] \leq e^{- t^2/2} \:.
\end{aligned}
\label{eq:prv3}
\end{eqnarray}
Thus, (\ref{eq:prv1}), (\ref{eq:prv2}) and (\ref{eq:prv3}) prove Lemma \ref{lm:perm_neg}.
\\
\\
\noindent
\textbf{Proof of Lemma \ref{lm:mylemma}}\\
First, we add and subtract the mean of each random component as follows
\begin{eqnarray}
\begin{aligned}
\underset{\|\bu\| = 1}{\sup} \:\: \sum_{i = 1}^{n} (\bu^T \bg_i)^2 &\leq \underset{\|\bu\| = 1}{\sup} \:\: \sum_{i = 1}^{n} \left[ (\bu^T \bg_i)^2 -  \mathbb{E} (\bu^T \bg_i)^2  \right] \\
& + \underset{\|\bu\| = 1}{\sup} \:\: \sum_{i = 1}^{n} \mathbb{E} (\bu^T \bg_i)^2.
\label{eq:main_1nq}
\end{aligned}
\end{eqnarray}
Similar to (\ref{eq:expectequal}),
\begin{eqnarray}
\underset{\|\bu\| = 1}{\sup} \:\: \sum_{i = 1}^{n} \mathbb{E} (\bu^T \bg_i)^2 =\frac{n}{N}  \:.
\label{mean_mylemaj}
\end{eqnarray}
The first component of the RHS of (\ref{eq:main_1nq}) can be rewritten as
\begin{eqnarray}
\begin{aligned}
 & \underset{\|\bu\| = 1}{\sup} \:\: \sum_{i = 1}^{n} \left[ (\bu^T \bg_i)^2 -  \mathbb{E} (\bu^T \bg_i)^2  \right]  \\
 & = \underset{\|\bu\| = 1}{\sup} \:\:
 \bu^T  \left( \sum_{i = 1}^{n} \bg_i \bg_i^T - \mathbb{E} \: \{ \bg_i \bg_i^T \}     \right) \bu \\
 & = \underset{\|\bu\| = 1}{\sup} \:\:
 \bu^T  \left( \sum_{i = 1}^{n} \bg_i \bg_i^T - \frac{1}{N} \bI     \right) \bu.
\end{aligned}
\end{eqnarray}
The matrices $\{ \bg_i \bg_i^T - \frac{1}{N} \bI \}_{i = 1}^n$ are zero mean random matrices. Thus, we use the non-commutative Bernstein inequality to bound the spectral norm of the matrix  $\bM$ defined as
\begin{eqnarray}
\bM =  \sum_{i = 1}^{n} \left( \bg_i \bg_i^T - \frac{1}{N} \bI    \right).
\end{eqnarray}

\begin{lemma}[\text{Non-commutative Bernstein inequality \cite{lamport11}}]
Let $\bX_1 , \bX_2 , ... , \bX_L $ be independent zero-mean random matrices of dimension $d_1 \times d_2$. Suppose $\rho_k^2 = \max \{\| \mathbb{E} [ \bX_k \bX_k^T ] \| , \| \mathbb{E} [ \bX_k^T \bX_k ] \| \} $ and $\| \bX_k \| \leq M$ almost surely for all k. Then for any $\tau > 0$
\begin{eqnarray}
\begin{aligned}
&\mathbb{P} \left[ \Bigg \| \sum_{k=1}^{L} \bX_k \Bigg  \| > \tau \right]  \\
& \leq (d_1 + d_2) \exp \left( \frac{-\tau^2 / 2}{\sum_{k=1}^L \rho_k^2  + M\tau/3} \right) .
\end{aligned}
\label{eq28}
\end{eqnarray}
\label{lm5}
\end{lemma}

\noindent
To find the parameter $M$ defined in Lemma \ref{lm5}, we compute
\begin{eqnarray}
\begin{aligned}
 & \| \bg_i \bg_i^T - \frac{1}{N} \bI \| \leq \max ( \| \bg_i \bg_i^T \| , \frac{1}{N} ) = 1
\end{aligned}
\end{eqnarray}
where we used the fact that $ \| \bH_1 - \bH_2  \| \leq \max ( \|\bH_1 \| , \|\bH_2 \|)$, if $\bH_1$ and $\bH_2$ are positive definite matrices.
Thus, $M = 1$. Similarly, for the parameter $\rho$ we have
\begin{eqnarray}
\begin{aligned}
 & \left\| \mathbb{E} \: \left[  \left( \bg_i \bg_i^T - \frac{1}{N} \bI \right) \left( \bg_i \bg_i^T - \frac{1}{N} \bI \right) \right] \right\| = \\
 &  \left\| \mathbb{E} \: \left[  \bg_i \bg_i^T - \frac{2}{N} \bg_i \bg_i^T  + \frac{1}{N^2} \bI \right] \right\| = \\
& \left\| \mathbb{E} \: \left[   \frac{1}{N^2} \bI - \frac{1}{N} \bI \right] \right\| \leq  \max(\frac{1}{N} , \frac{1}{N^2}) = \frac{1}{N}.
\end{aligned}
\end{eqnarray}
Therefore, according to Lemma \ref{lm5},
\begin{eqnarray}
\begin{aligned}
&\mathbb{P} \left[ \| \bM \| > \tau \right] \leq
 2 N \exp \left( \frac{-\tau^2 / 2}{n/N  +  \tau/3} \right) .
\end{aligned}
\end{eqnarray}
Thus,
\begin{eqnarray}
\begin{aligned}
&\mathbb{P} \left[ \| \bM \| > \eta \right] \leq
\delta
\end{aligned}
\label{eq:uperbound_mylm}
\end{eqnarray}
where
\begin{eqnarray}
 \eta = \max \left( \frac{4}{3} \log \frac{2N}{\delta} , \sqrt{4 \frac{n}{N} \log \frac{2 N}{\delta}} \right) \: .
\label{eta_def}
\end{eqnarray}

According to (\ref{mean_mylemaj}) and (\ref{eq:uperbound_mylm}),
\begin{eqnarray}
\begin{aligned}
& \underset{\|\bu\| = 1}{\sup} \:\: \sum_{i = 1}^{n} (\bu^T \bg_i)^2 <  \frac{n}{N} + \eta
\end{aligned}
\end{eqnarray}
with probability at least $1 - \delta$, where $\eta$ is defined in (\ref{eta_def}).
\\
\\
\noindent
\textbf{Proof of Lemma \ref{lm:mylemman}}\\
The proof of this Lemma is similar to the proof of Lemma \ref{lm:mylemma}. First, we add and subtract the mean values to expand the LHS of (\ref{eq:suff_perm_2n}) as follows
\begin{eqnarray}
\begin{aligned}
& \underset{\|\bu\| = 1}{\inf} \:\: \sum_{i = 1}^{n} ( \bu^T \bg_i )^2 \ge \\
& \underset{\|\bu\| = 1}{\inf} \:\: \sum_{i = 1}^{n} \left[ ( \bu^T \bg_i )^2 - \mathbb{E} ( \bu^T \bg_i )^2 \right] + \underset{\|\bu\| = 1}{\inf} \:\: \sum_{i = 1}^{n} \mathbb{E} ( \bu^T \bg_i )^2.
\end{aligned}
\label{mean_myleman}
\end{eqnarray}
Similar to (\ref{mean_mylemaj}),
\begin{eqnarray}
\underset{\|\bu\| = 1}{\inf} \:\: \sum_{i = 1}^{n} \mathbb{E} ( \bu^T \bg_i )^2 =  {\frac{n}{N}} \:.
\label{eq:prv2n}
\end{eqnarray}
Based on the analysis presented in the proof of Lemma \ref{lm:mylemma}, we can conclude that
\begin{eqnarray}
 & \underset{\|\bu\| = 1}{\sup} \:\: \sum_{i = 1}^{n} \left[ \mathbb{E} (\bu^T \bg_i)^2 -  (\bu^T \bg_i)^2   \right] < \eta
\end{eqnarray}
with probability at least $1- \delta$, where $\eta$ is given in (\ref{eta_def}).
Hence,
\begin{eqnarray}
& \mathbb{P} \left[ \underset{\|\bu\| = 1}{\inf} \:\: \sum_{i = 1}^{n} ( \bu^T \bg_i )^2 \leq \frac{n}{N} - \eta \right] < \delta \:.
\end{eqnarray}

%\nocite{*}
%\bibliographystyle{IEEEbib}
%%{\small
%\bibliography{bibfile}%}

\bibliographystyle{IEEEtran}
\bibliography{IEEEabrv,bibfile}

\end{document}